\documentclass[runningheads]{llncs}
\usepackage[T1]{fontenc}
\usepackage{graphicx}
\usepackage{marvosym}
\usepackage{amssymb}   
\usepackage{amsmath}   
\usepackage{thmtools,thm-restate}
\usepackage{enumitem}
\usepackage{dsfont}
\usepackage{upgreek}
\usepackage{hyphenat} 
\usepackage[dvipsnames,table]{xcolor}
\usepackage{pifont} 
\usepackage{multirow}
\usepackage{booktabs} 
\usepackage{setspace}              
\usepackage{caption}
\usepackage{makecell}  
\usepackage[ruled,linesnumbered]{algorithm2e}
\usepackage{mathrsfs} 
\usepackage{wrapfig} 
\usepackage{hyperref}
\usepackage{float}
\usepackage{layouts}
\usepackage{placeins} 

\newcommand{\indicator}{\mathds{1}}

\newcommand{\methodName}{\textbf{OCULAR}}
\newcommand{\methodNameLong}{\textbf{O}bservation-aware \textbf{C}onformal \textbf{U}ncertainty \textbf{L}ocal-C\textbf{a}lib\textbf{r}ation}
\newcommand{\position}{p_{t}}
\newcommand{\velocity}{v_{t}}

\newcommand{\state}{s_{t}}
\newcommand{\stateSpace}{\mathcal S}

\newcommand{\observation}{o_{t}}
\newcommand{\observationCone}{o^\prime_{t}}
\newcommand{\projectingFunc}[1]{\textsc{Project}\textup{(}#1\textup{)}}

\newcommand{\encoder}{\textsc{Encode}}

\newcommand{\latent}{\textsc{latent}}
\newcommand{\latentSpace}{\mathbb{R}^{\dim(\latent)}}
\newcommand{\labelSet}{\mathcal L}

\newcommand{\processFunc}[1]{\textsc{process}\textup{(}#1\textup{)}}
\newcommand{\action}{a_t}
\newcommand{\actionSpace}{\mathcal A}
\newcommand{\trueDynamics}{f}
\newcommand{\approxDynamics}{\tilde{f}}
\newcommand{\disturbance}{w_t}
\newcommand{\mismatch}{\boldsymbol{\color{orange}{1.3}}}
\newcommand{\failureRate}{\upalpha}
\newcommand{\CPinput}{X}
\newcommand{\CPinputTest}{X_{n+1}}
\newcommand{\CPinputSpace}{\mathcal X}
\newcommand{\CPoutput}{Y}
\newcommand{\CPoutputTest}{Y_{n+1}}
\newcommand{\CPoutputSpace}{\mathcal Y}

\newcommand{\scoreFunc}{r}
\newcommand{\scoreVal}{R}

\newcommand{\calibset}{D_{\mathrm{cal}}}

\newcommand{\numcal}{n}
\newcommand{\calibDynamics}{\tilde{f}_{cal}}
\newcommand{\CPregion}{\hat{\mathcal C}}

\newcommand{\numSemanticClass}{M}

\newcommand{\predSpace}{\mathcal H}

\spnewtheorem{assumption}{Assumption}{\bfseries}{\itshape}

\newif\ifarxiv
\arxivtrue 

\ifarxiv

\newcommand{\ccell}[2]{%
  \cellcolor{#1}\rule{0pt}{2.4ex}#2\rule[-1.2ex]{0pt}{0pt}}
\newcommand{\mycheck}{{\color{ForestGreen}\ding{51}}}
\newcommand{\myxmark}{{\color{red}\ding{55}}}



\else
\hypersetup{hidelinks}
\newcommand{\ccell}[2]{%
  \rule{0pt}{2.4ex}#2\rule[-1.2ex]{0pt}{0pt}}
\newcommand{\mycheck}{\ding{51}}
\newcommand{\myxmark}{\ding{55}}



\fi

\begin{document}
\title{
Local Conformal Calibration of Dynamics Uncertainty from Semantic Images}
\titlerunning{Local Conformal Calibration of Dynamics Uncertainty from Semantic Images}
%
\author{Lu\'is Marques\textsuperscript{(\Letter)}\orcidID{0000-0001-7211-411X} \and
Dmitry Berenson\orcidID{0000-0002-9712-109X}}
\authorrunning{L. Marques and D. Berenson}
%
\institute{Robotics Department, University of Michigan, Ann Arbor, MI 48109, USA
\\
\email{\{lmarques,dmitryb\}@umich.edu}}
\maketitle              
\begin{abstract}
We introduce \textbf{O}bservation-aware \textbf{C}onformal \textbf{U}ncertainty \newline \textbf{L}ocal-C\textbf{a}lib\textbf{r}ation (\methodName), a conformal prediction-based algorithm that uses perception information to provide \textit{uncertainty quantification guarantees} for \textit{unseen} test-time environments.
While previous conformal approaches lack the ability to discriminate between state-action space regions leading to higher or lower model mismatch and require environment-specific data, our method uses data collected from visually similar environments to provably calibrate a linear Gaussian dynamics model of arbitrary fidelity.
The prediction regions generated from \methodName~are guaranteed to contain the future system states with, at least, a user-set likelihood, despite both aleatoric and epistemic uncertainty---i.e., uncertainty arising from both stochastic disturbances and lack of data. 
Our guarantees are non-asymptotic and \textit{distribution-free}, not requiring strong assumptions about the \textit{unknown} real system dynamics.
Our calibration procedure enables distinguishing between observation-velocity-action inputs leading to higher and lower next-state-uncertainty, which is helpful for probabilistically safe planning.
We numerically validate our algorithm on a double-integrator system subject to random perturbations and significant model mismatch, using both a simplified sensor and a more realistic simulated camera.
Our approach calibrates approximate uncertainty estimates both when in-distribution and out-of-distribution, producing \textit{volume-efficient} prediction regions without requiring environment-specific data.

$\quad$Project website: \href{https://um-arm-lab.github.io/ocular}{\texttt{https://um-arm-lab.github.io/ocular}}\par

\end{abstract}
\section{Introduction}
\vspace{-3mm}
\looseness-1 Provable uncertainty quantification typically requires imposing restrictive assumptions on the distribution of external disturbances \cite{5970128}
or assuming a lack of model mismatch \cite{devCDC21}.
However, real systems are often subject to disturbances that are difficult to model analytically (e.g., near-wall aerodynamic effects) and are action-dependent (e.g., tire slip on ice can vary with velocity).
The discrepancy between the strict conditions required for classical guarantees and realistic deployment conditions can lead to unsafe actions.
Conversely, while learned dynamical systems have made progress towards modeling complex dynamics \cite{shi19_lander}, their predictions are unreliable when operating outside the training regime.
It remains challenging to provide theory-grounded guarantees on the uncertainty present in next-state dynamics prediction when subject to unknown disturbances and when relying on a dynamics model of unknown fidelity.

\looseness-1 Conformal Prediction (CP) has recently emerged as a statistical framework for \textit{distribution-free} uncertainty quantification. CP does not limit the form of the unknown dynamics or the external disturbances.
CP provides \textit{provable probabilistic guarantees} given a finite amount of \textit{calibration data}, enabling a data-driven but theoretically-grounded quantification of prediction uncertainty.
However, CP still lacks some properties limiting its applicability in robotics.
The CP guarantees are conventionally given on average over a set of possible conditions and require system-transition data representative of test-conditions.
Hence, calibration data is conventionally 
collected from the same environment as test-time deployment \cite{lindemann2023safe,lucca,claps}---a significant shortcoming.
Besides
poor generalizability, per-environment data-collection can expose the system to unnecessary risks.
For example, if the CP guarantees depend on robot poses, then we might have to ``completely'' traverse a new road to determine in which of its segments our dynamics model is more or less accurate (e.g., which parts are icy).
However, if the guarantees depend instead on perception information (e.g., a semantic image of the pavement ahead of the vehicle), then we might be able to quantify uncertainty in scenarios that have not yet been tested, but that are \textit{visually-similar} to previous examples.
That is, we could possibly learn that going fast on icy roads might lead to high dynamics uncertainty, without having to test all icy roads independently.
In this paper we tackle this limitation by introducing a CP-based algorithm that utilizes robot-frame perception information to achieve the same coverage guarantees as existing approaches \textit{without requiring any calibration data from the test environment}.
Instead, we utilize data from different, but perceptually similar environments, and a learned representation of observations to safely generalize to unseen situations.
Furthermore, we account for how uncertainty might vary across velocity-action-observation contexts and provide an adaptive calibration of approximate Gaussian uncertainty estimates, enabling the construction of a prediction region that will contain the unknown next state with a user-set likelihood of $(1-\failureRate)\in(0,1)$.
Our key contributions are
\vspace{-4mm}
\begin{enumerate}[label=\roman*)]
    \item We propose an algorithm that, given state, action, and perception information from visually-similar environments, provides provable finite-sample guarantees on dynamics uncertainty in a new unseen environment. 
    \item We prove the validity of our approach, which combines learned representations with local conformal calibration, and demonstrate its use for constructing safe plans that steer the system towards lower-uncertainty states.
    \item We validate our approach on a double-integrator system using both planar-sensors and Isaac Sim cameras, demonstrating comparable safety and performance to methods requiring environment-specific data.\footnote{See \href{https://um-arm-lab.github.io/ocular}{{https://um-arm-lab.github.io/ocular}} for planning videos of all methods.}
\end{enumerate}
\vspace{-5mm}
\section{Related Work}
\vspace{-3mm}
\looseness-1 \textbf{Safe Planning under Uncertainty.} This problem setting has a long history, with approaches including tube-based MPC \cite{jeantube}, contingency-based MPC \cite{8815260}, using Gaussian processes to quantify aleatoric uncertainty \cite{khan2021safety}, and ensemble-based models to quantify epistemic uncertainty \cite{wang2024providing}. Classical approaches, such as control barrier functions \cite{ames2019control} or reachability-based planning \cite{bansal2017hamilton}, may be computationally expensive online or require a certain system structure. Conversely, learning-based methods often provide looser uncertainty estimates, especially when out-of-distribution (OOD). In this work, we use a data-driven and theoretically-grounded approach to provide probabilistic guarantees on one-step safety and a volume-efficient calibration of approximate Gaussian uncertainty models.

\textbf{Conformal Prediction in Robotics.} CP has been used for socially-aware robot navigation \cite{lindemann2023safe,pmlr-v211-dixit23a}, expert imitation \cite{sun2023conformal}, failure detection \cite{luo2024sample}, synthesis of control barrier functions \cite{zhou2024safety}, and calibration of Lie-algebraic uncertainty estimators \cite{claps}. However, existing methods perform a state-, action-, and observation-agnostic calibration of system uncertainty.
Yet, for some robot systems and approximate dynamics models, the uncertainty underlying a given state-transition may not be constant. Empirically, global calibration methods have limited utility for safe plan generation, as these cannot steer the system towards lower-uncertainty states.
While \cite{lucca} proposed a position-velocity-action-dependent calibration, its reliance on world-frame poses requires collecting data on each deployment environment.
In contrast, we leverage perception information to enable local calibration of dynamics uncertainty in \textit{environments for which we have not collected any data}. 
While Online CP methods \cite{pmlr-v211-dixit23a,ACP} do not assume a calibration dataset, these provide input-independent asymptotic guarantees. Instead, our velocity-action-observation-dependent guarantees are valid for finitely-sized datasets collected in different environments from the test environment.

\textbf{Conformal Prediction with Sensing.} CP has been used as a wrapper on vision-based state-estimators, to probabilistically bound unmodeled sensing uncertainty and lead to theory-grounded state estimates \cite{stateEstimateCP,VOestimateCP}. 
In \cite{mei2024perceive}, bounding-box predictors and occupancy predictors were calibrated to build regions containing the true obstacles at a specified likelihood.
While useful for provably safe-autonomy, these works perform an input-independent calibration of learned sensing modules. We do not focus on state or obstacle estimation, assuming the current robot pose and velocity to be known, and that obstacles can be adequately observed. Rather, we provide an observation-dependent local calibration of system transition uncertainty, which we demonstrate can lead to more efficient and safe plans than CP baselines performing global calibration.

\vspace{-5mm}
\section{Problem Statement}\label{sec:probStatement}
\vspace{-4mm}
\looseness-1 Let $\state\in \stateSpace$ and $\action \in \actionSpace$ denote a robot's state and action respectively, where $t\in \mathbb N_0$.
We consider discrete-time \textit{stochastic} \textit{dynamical} systems evolving according to some \textit{unknown} time-invariant process $(\state{}_{+1}) \sim \trueDynamics (\state, \action)$.
We assume the state can be decomposed into a pose element $\position$, defined relative to an inertial frame, and a body-velocity element $\velocity$, so that $\state := (\position, \velocity)$. We further assume the robot is equipped with a depth sensor and an algorithm providing semantically segmented images (e.g., \cite{ren2024grounded,simeoni2025dinov3}) where each pixel has an associated class.
We consider as observation $\observation :=(\observation^{depth},\observation^{semantics})$, where $\observation=h(\state)$ for some $h$.
The randomness inherent in $\trueDynamics$, termed \textit{aleatoric uncertainty}, can arise, for example, due to unmodeled disturbances (e.g., wind gusts, uneven terrain) and wheel slip. We assume that this uncertainty cannot be reduced by gathering more data. 
\looseness-1
Without restricting the form or the noise distribution of the \textit{unknown} dynamics $\trueDynamics$, we assume access to an approximate probabilistic dynamics model $\approxDynamics$ of unknown fidelity.
In this work, $\approxDynamics$ is a linear Gaussian dynamics model whose predictions are multivariate normals capturing a distribution over the next state $\approxDynamics : \tilde{\mathcal{N}}_t, \action \mapsto \tilde{\mathcal{N}}_{t+1}$.
Model mismatch will introduce
\textit{epistemic uncertainty}, e.g., if $\trueDynamics$ is multimodal or heavy-tailed.
Likewise, epistemic uncertainty may increase when making predictions in regions of the state-action space that are poorly covered by the training data.
For example, consider a car-like dynamics model $\approxDynamics$ that was tuned for nominal driving conditions and is then deployed on an icy road.
The lower test-time ground friction might lead to $\state{}_{+1}$ having greater variability than what $\approxDynamics$ estimates.
Additionally, the amount of epistemic uncertainty can be heterogeneous across the state-action space.
In principle, epistemic uncertainty can be reduced by further data collection and model updates, but doing so \textit{safely} is nontrivial. 
We therefore consider the setting where $\approxDynamics$ is fixed. Without making strong assumptions about $\trueDynamics$, it remains difficult to accurately quantify the amount of aleatoric and epistemic uncertainty present. \newline\indent
We require \textit{calibrated} prediction regions\footnote{See 
\ifarxiv
App. \ref{app:extramotivation}
\else
\cite{wafr26arxiv}, App. A,
\fi
for some consequences of using \textit{uncalibrated uncertainty predictions}.
} $\CPregion$ with \textit{frequentist validity}, i.e., regions whose \textit{marginal coverage}\footnote{The \textit{coverage} of a prediction region is the probability that it will contain the random variable of interest, in our case the true next robot state $\state{}_{+1}$.} is at least their confidence level of $(1-\failureRate)$. Mathematically, this can be expressed as the requirement 
\vspace{-2mm}
\begin{equation}\label{eq:marginalCoverage}
    \mathbb P\{ \CPoutputTest \in \CPregion(\CPinputTest)\} \ge (1-\failureRate),
\vspace{-2mm}
\end{equation}
\looseness-1 where for convenience we defined $\CPoutputTest:=\state{}_{+1}$ and $\CPinputTest$ is a CP input, e.g., the car's position $\position$, its velocity $\velocity$, or the driver's action $\action$.
Purely achieving Eq \eqref{eq:marginalCoverage} is trivially done with $\CPregion(\CPinput)=\CPoutputSpace,\forall \CPinput$. Yet, this is hardly informative or helpful.
For our setting, such a region would cover the entire state-action space, which would be useless for planning and control. 
Hence, we want our prediction regions to also be \textit{volume-efficient}, i.e., as small as possible, and \textit{adaptive}---smaller for lower-uncertainty transitions and larger for higher-uncertainty transitions.

Our objective is then three-fold. First, we want an algorithm that can \textit{provably calibrate} approximate models $\approxDynamics$ to make their prediction regions valid in a frequentist sense at a user-set acceptable failure rate $\failureRate\in (0,1)$, \textit{without imposing assumptions about the distribution} of the uncertainty.
Second, we aim to use the calibrated model $\calibDynamics$ for \textit{probabilistically safe planning} in environments where $\approxDynamics$ might lead to collisions.
This requires $\CPregion$ to be \textit{adaptive}, and \textit{volume-efficient}.
Thirdly, we want the calibration procedure to have some level of generalizability and data-efficiency.
To make these objectives tractable, we assume access to a dataset $\calibset$ of robot-frame state-transitions $(\state,\action,\observation, \state{}_{+1})$ executed in environments \textit{different from}, but \textit{visually-similar to}, the deployment environment.
We use said transitions to compare different $\approxDynamics$ predictions with the realized resulting states $\state{}_{+1}$.
We also assume that robot dynamics are consistent across environments (e.g., ice is equally slippery across maps).
Formally:
\begin{assumption}\label{ass:exchangeable}
    We have a dataset of transitions $\calibset := \{(\state, \action, \observation, \state{}_{+1})_i\}_{i=1}^{n}$ that is exchangeable\footnote{The random variables $\calibset \cup \{(\state, \action, \observation, \state{}_{+1})_{n+1}\}$ are \textit{exchangeable} if they are equally likely to appear in any ordering. This is weaker than assuming that they are i.i.d.
    }
    with the test-time transitions $(\state, \action, \observation, \state{}_{+1})_{n+1}$.
\end{assumption}
Exchangeability implies that test-time transitions must not be OOD relative to the calibration dataset, i.e., we should have traversed over some ice before being able to determine that icy roads might increase predictive uncertainty.
Yet, they might be OOD relative to the original approximate model $\approxDynamics$.
Following standard CP literature \cite{lindemann2023safe,pmlr-v211-dixit23a,lucca,claps}, the data in $\calibset$ is uncorrupted.
\vspace{-5mm}
\section{Theoretical Background on Conformal Prediction}\label{sec:backgroundCP}
\vspace{-3mm}
Below we briefly introduce necessary concepts from the conformal prediction (CP) literature and refer the interested reader to \cite{angelopoulos2025theoreticalfoundationsconformalprediction,vovk2005algorithmic} for deeper study.
Following this literature, the $(\cdot )_{n+1}$ subscript refers to test-time transitions, so $\CPoutput_{1:n}$ are the observed $\state{}_{+1} \in \calibset$ and $\CPoutputTest$ is the unobserved $\state{}_{+1}$ found at inference time. For now, consider the prediction region input $\CPinput_{i}$ to be some function of $(\state,\action,\observation)_{i}$, for $i\in\{1,\ldots n+1\}$, that preserves exchangeability between calibration and test points, i.e., $\calibset:=\{(\CPinput,\CPoutput)_i\}_{i=1}^{\numcal}$ is still exchangeable with $(\CPinput,\CPoutput)_{n+1}$. Section \ref{sec:methodSection} describes different design-choices for $\CPinput$ that preserve the exchangeability of Assumption \ref{ass:exchangeable}.
Likewise, the Gaussian model predictions can be written as $\approxDynamics(\CPinput) \in \predSpace$, where the prediction space $\predSpace$ contains the possible state-space multivariate normals $\tilde {\mathcal N}_{t+1}$.

\vspace{-4mm}
\subsection{Split Conformal Prediction}\label{sec:SplitCP}
\vspace{-3mm}
The overall goal of CP is to construct a prediction region $\CPregion(\CPinputTest)$ given an observed test-time input $\CPinputTest$, that contains the unobserved test-time output $\CPoutputTest$ with (at least) a user-defined likelihood of $(1-\failureRate)$, i.e., prediction regions that satisfy Eq \eqref{eq:marginalCoverage}.
Split Conformal Prediction (SplitCP) is a variant widely used in robotics \cite{lindemann2023safe,claps,VOestimateCP,stateEstimateCP} due to its computational speed and minimal assumptions on the true uncertainty distribution and $\approxDynamics$.

Let $\scoreFunc : \mathcal \predSpace \times \CPoutputSpace \to \mathbb R$ be a user-defined scalar-valued \textit{nonconformity score} that measures disagreement between model predictions and true labels, with lower values corresponding to more accurate predictions.
Looping over every element in $\calibset$, we can then compute a measure of how accurate $\approxDynamics$ was for each transition, i.e., calculate $\scoreVal_i := \scoreFunc (\approxDynamics(\CPinput_i^{raw}), \CPoutput_i)$.
The test-time score $\scoreVal_{n+1}$ is unknown since $\CPoutputTest$ is unknown, so we conservatively set $\scoreVal_{n+1}=\infty$.
Since $\approxDynamics$ and $\scoreFunc$ are fixed before observing $\calibset$ or ${(\CPinput,\CPoutput)_{\numcal+1}}$, then the scores $\{\scoreVal_i \}_{i=1}^{\numcal+1}$ are still exchangeable \cite{angelopoulos2025theoreticalfoundationsconformalprediction,vovk2005algorithmic}.
Using a rank statistics argument \cite{angelopoulos2025theoreticalfoundationsconformalprediction,vovk2005algorithmic}, the rank of $\scoreVal_{n+1}$ can be shown to be uniformly distributed over $\{1,\ldots n,n+1\}$. Hence, it follows that $\mathbb P\{\scoreVal_{n+1}\le \hat{q} \}\ge(1-\failureRate)$ where $\hat{q} \in \mathbb R$ is the $(1-\failureRate)$ quantile of $\{\scoreVal_i\}_{i=1}^{n+1}$ \cite{angelopoulos2025theoreticalfoundationsconformalprediction,vovk2005algorithmic}.
Intuitively, we now have a probabilistic upper bound $(\hat{q})$ for the test-time nonconformity score $\scoreVal_{n+1}$, that holds (at least) at the user-specified likelihood of $(1-\failureRate)$.
This upper bound $(\hat{q})$ will take the value of one of the scores observed in the calibration dataset
and can be determined \textit{offline}.
Then, by construction, the prediction region
\vspace{-2mm}
\begin{equation}\label{eq:cpRegionConstruction}
    \CPregion(\CPinputTest) := \{y\in \CPoutputSpace: \scoreFunc(\approxDynamics(\CPinputTest),y)\le \hat{q} \}\vspace{-2mm}
\end{equation}
provides \textit{marginal coverage guarantees}, i.e., satisfies Eq \eqref{eq:marginalCoverage}.
If $\scoreFunc$ is the Euclidean norm, then $\CPregion$ will be ball-shaped, and if $\scoreFunc$ is the Mahalanobis distance between the predictive Gaussian $\approxDynamics(\CPinput)$ and the true label $\CPoutput$, then $\CPregion$ will be a hyperellipsoid.
While the above marginal coverage guarantees hold for (practically) all $\scoreFunc$, some nonconformity scores will lead to more efficient or adaptive algorithms than others. Further, the score upper bound $\hat{q}$ is invariant to $\CPinput$, while it is clear that $\approxDynamics$'s uncertainty might be heterogeneous across the state-action space. Ideally, $\hat{q}$ would be $\CPinput$-dependent so that a planner could guide motions towards regions of the state-action space with lower uncertainty (e.g., by penalizing higher $\hat{q}$).
Ultimately, \textit{marginal coverage guarantees} hold on average, over the distribution of $\calibset$ and test cases, and not for a specific test transition. 
\vspace{-3mm}
\subsection{Local Conformal Prediction}\label{sec:localCP}
\vspace{-2mm}
For robotics, it would be desirable instead to construct regions with \textit{(input-) conditional coverage}, i.e., satisfying
$\mathbb P\{ \CPoutputTest \in \CPregion \mid \CPinputTest = x\} \ge (1-\failureRate)$.
This approach would give probabilistic guarantees for each specific test-time transition.
Unfortunately, it was proven that this is impossible for finitely-sized $\calibset$ \cite{lei2014distribution}, if we do not want the trivial region $\CPregion=\CPoutputSpace$.
Instead, we seek a compromise between the ideal conditional coverage and the standard marginal coverage.
Our algorithm will provide \textit{finite-sample local coverage} (LocalCP) \cite{lei2014distribution}.
Given a predetermined disjoint partition of the input space $\CPinputSpace = \sqcup_{k=1}^K \CPinputSpace_k$, our prediction regions will satisfy
\vspace{-3mm}
\begin{equation}
    \mathbb P\{ \CPoutputTest \in \CPregion(\CPinputTest) \mid \CPinputTest \in \CPinputSpace_k\} \ge (1-\failureRate),\forall k \in \{1,\ldots, K \},
    \vspace{-3mm}
\end{equation}
\looseness-1 
where the probability is now taken over each partition \cite{lei2014distribution}.
While SplitCP could severely undercover for regions of $\CPinputTest$ that were underrepresented in $\calibset$, LocalCP provides per-partition guarantees. 
Additionally, each partition $\CPinputSpace_k$ will have its own uncertainty threshold $\hat{q}_k \in \mathbb R$, enabling disambiguation between high and low uncertainty states.
In practice, the partitioning process can significantly impact the generalizability and the empirical performance of LocalCP.
Our algorithm learns a partitioning scheme using $(\velocity,\action,\observation)$.

\vspace{-5mm}
\section{Method: \methodName}\label{sec:methodSection}
\begin{figure}[t]
\includegraphics[width=\textwidth]{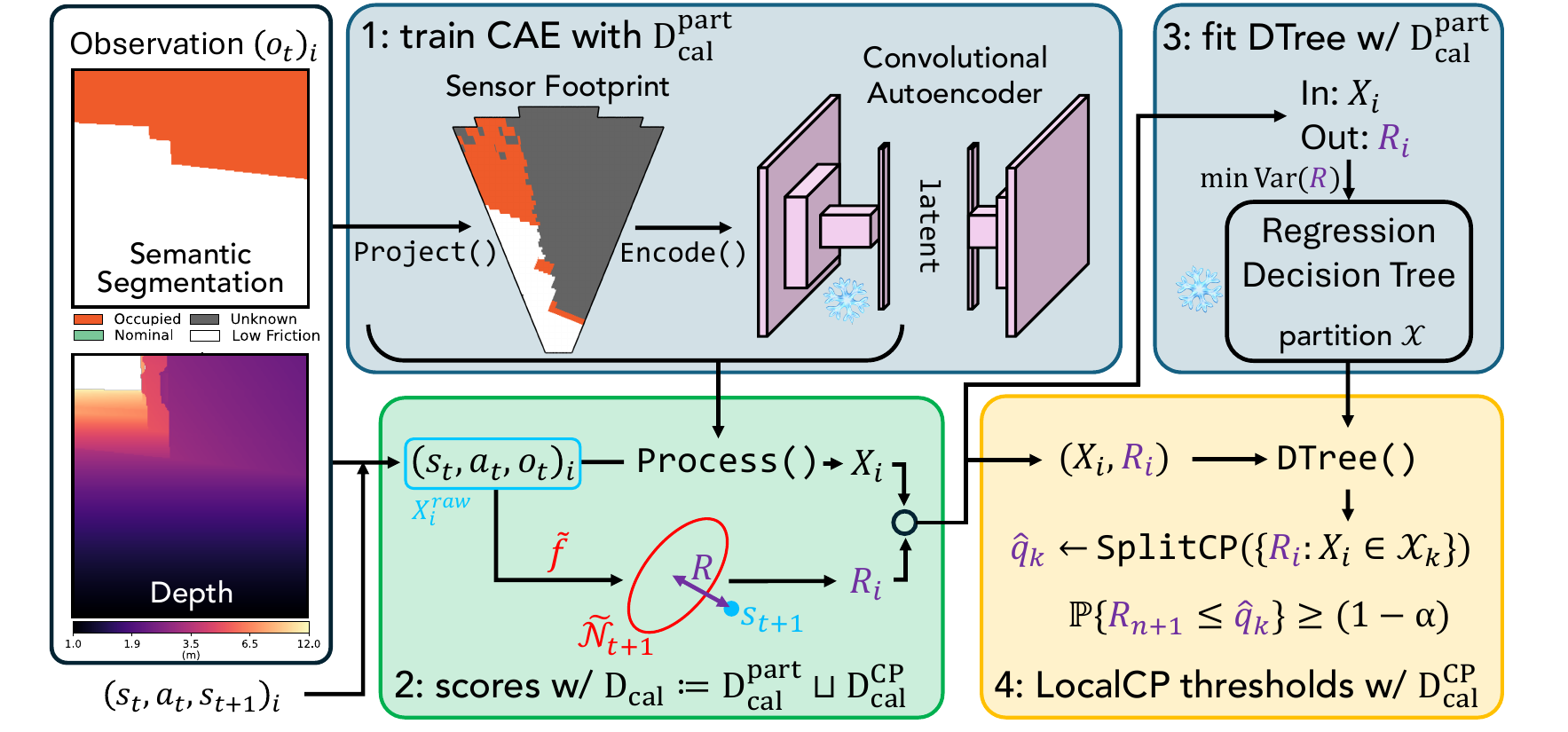}\vspace{-3mm}
\caption{Offline component of \methodName. 1: $\observation$ from $\calibset^{part}$ is projected into a planar footprint $\observationCone$ by $\projectingFunc{}$. A CAE is trained to reconstruct $\observationCone$, and the decoder is discarded. 2: All data in $\calibset$ is processed by $\processFunc{}$ into a learned representation $\CPinput_i$, and nonconformity scores $\scoreVal_i$ are computed. 3: a Decision Tree is trained on $\calibset^{part}$ to partition the learned input space $\CPinputSpace$ into regions of approximately constant score. 4: The holdout processed $\calibset^{CP}$ data is fed through the DTree and scores are grouped per leaf node $k$. SplitCP is performed on each input-space partition $\CPinputSpace_k$ to get an input-dependent probabilistic threshold $\hat{q}_k$ on test-time scores $R_{n+1}$.} \label{fig:offlineMethod}\vspace{-6mm}
\end{figure}

\begin{figure}[t]
\includegraphics[width=\textwidth]{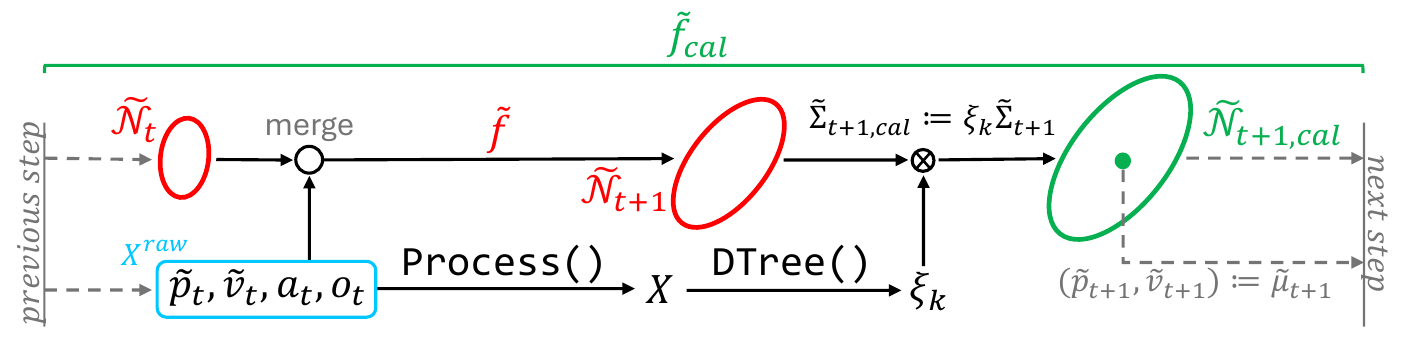}\vspace{-4mm}
\caption{\looseness-1 Online component of \methodName~$(\approxDynamics_{cal})$. Given an estimated Gaussian at time $t$, a desired action $\action$, and observation $\observation$, we create an approximate next-step Gaussian $\tilde{\mathcal N}_{t+1}$ via the approximate model $\approxDynamics$. The current-time information $\CPinput^{raw}$ is processed, and the learned representation $\CPinput$ is passed to the Decision Tree. The resulting leaf node $\CPinputSpace_k$ has an associated $\hat{q}_k$
which is multiplied by a fixed constant to get
$\xi_k$. $\tilde{\mathcal N}_{t+1}$ is then calibrated by scaling its covariance by $\xi_k$ and the output is passed to the following planning step.} \label{fig:onlineMethod} \vspace{-8mm}
\end{figure}

\vspace{-3mm}
\looseness-1 We introduce \methodNameLong \newline(\methodName), a method that provably calibrates the approximate uncertainty estimates of a linear Gaussian dynamics model through the use of a finite-sized calibration dataset $\calibset$ of robot-frame data (depth and semantic observations, actions, and body-frame velocities).
By leveraging robot-frame camera observations instead of inertial-frame poses, our approach does not require calibration data in the test environment (a major limitation of previous approaches \cite{lindemann2023safe,lucca,claps}). Instead, it generalizes to environments which are visually-similar to the data in $\calibset$.
First, we give an overview of how LocalCP is applied in robotics, and an intuition for the impact of different input-space $\CPinputSpace$ choices (Sec.~\ref{sec:methodLocal}). Then, we detail how we convert observations into a canonical form $\observationCone$, and subsequently encode them into a low-dimensional latent representation $\latent$ (Sec.~\ref{sec:obsprocessing}). 
Next, we prove that these transformations preserve exchangeability and combine our pipeline with LocalCP to achieve \textit{finite-sample local coverage} (Sec.~\ref{sec:methodexchange}).
Finally, we demonstrate how the \textit{calibrated dynamics model} $\calibDynamics$ resulting from our algorithm can be used for safe planning (Sec.~\ref{sec:methodPlanning}).
\methodName's offline component is depicted in Fig. \ref{fig:offlineMethod}, the online component in Fig. \ref{fig:onlineMethod}, and the full method is described in Alg.~\ref{alg:ocular}.

\vspace{-3mm}
\subsection{Steps for Local Calibration}\label{sec:methodLocal}
\vspace{-2mm}
\looseness-1 While LocalCP provides partition-level guarantees, it is crucial to construct partitions $\CPinputSpace_k$ that separate regions where $\approxDynamics$ is highly uncertain and less so.
While the partitioning is ultimately learned, we can construct representations that facilitate learning helpful $\CPinputSpace_k$.
Following standard LocalCP algorithms \cite{locart,lucca}, we divide $\calibset:= \calibset^{part} \sqcup \calibset^{CP}$ into a subset used to construct the partitioning scheme ($\calibset^{part}$) and another subset ($\calibset^{CP}$) to perform calibration on each $\CPinputSpace_k$, resulting in a per-$\CPinputSpace_k$ threshold $\hat{q}_k$.
This is required theoretically to preserve exchangeability between $\calibset^{CP}$ and test-time tuples \cite{locart}, which is key for our guarantees.

\looseness-1 From Ass. \ref{ass:exchangeable}, $\CPinput$, the input used to calibrate $\approxDynamics$'s uncertainty estimate of $\CPoutput:=\state{}_{+1}$, can be a function of our knowledge at time $t$, i.e., $\CPinput^{raw}:=(\state,\action,\observation)=(\position,\velocity,\action,\observation)$. LUCCa \cite{lucca} uses $X_i:= (\position,\velocity,\action)_i$, creating partitions in position-velocity-action space.
However, this design-choice requires collecting data in each new environment we want to traverse, due to the inclusion of the robot's location in world-coordinates ($\position$).
Instead, we drop $\position$ and create partitions in a learned velocity-action-latent-observation space via $X_i:=\processFunc{\CPinput^{raw}_i}$.
By considering the robot-frame inputs, we aim to improve generalization to new scenarios.
We now detail $\processFunc{}$.

\vspace{-5mm}
{%
\SetAlCapFnt{\small}
\SetAlCapNameFnt{\small}
\begin{algorithm}[!hb]
\caption{\small \methodNameLong}
\label{alg:ocular}
\SetInd{0.25em}{1em}
\SetAlgoLined
\DontPrintSemicolon

\KwIn{$\approxDynamics,\projectingFunc{},\calibset:=\calibset^{part}\sqcup\calibset^{CP},\failureRate,J, H,\texttt{Goal}$}

\BlankLine
\SetKwComment{Comment}{$//\ $}{}
\tcc{Define $X_i^{raw}:=(\position,\velocity,\action,\observation)_i$ and $Y_i:=(\state{}_{+1})_i$. See Sec.~\ref{\detokenize{sec:methodLocal}}.}
\vspace{-2pt}
\tcc{(Offline) on partition set $\calibset^{part}:=\{X^{raw}_i, Y_i\}_{i=1}^{\lvert \calibset^{part}\rvert}$. See Sec.~\ref{\detokenize{sec:obsprocessing}}.}
$(\observationCone)_i \leftarrow \projectingFunc{(\observation)_i},\ \ \forall\{(\observation)_i\}\in\calibset^{part}$
Train Convolutional AE on $\{(\observationCone)_i\}_{\in\calibset^{part}}$; keep $\encoder()$\;

Define $X_i:=\processFunc{\CPinput^{raw}_i}:=(\velocity,\action,\latent)_i$, where $\latent=\encoder(\projectingFunc{\observation)}$\;
\BlankLine
\tcc{(Offline) on the full calibration data $\calibset$. See Sec.~\ref{\detokenize{sec:methodexchange}}.}

$\CPinput_i \leftarrow \processFunc{\CPinput^{raw}_i}\ \ \forall\{\CPinput^{raw}_i\}\in\calibset$ \hfil \tcp*{\small{Process $\observation$, drop $\position$}}
$\scoreVal_i \leftarrow \scoreFunc(\approxDynamics(\CPinput^{raw}_i),\CPoutput_i)\ \ \forall\{\CPinput^{raw}_i,\CPoutput_i\}\in\calibset$\tcp*[f]{\small{Get nonconformity scores}}

\BlankLine
\tcc{(Offline) fit $\CPinputSpace_k$ on $\calibset^{part}$, calibrate on $\calibset^{CP}$. See Sec.~\ref{\detokenize{sec:methodexchange}}.}

$\texttt{DTree} \leftarrow \texttt{LOCART}(\{\CPinput_i,\scoreVal_i\}{\in\calibset^{part}})$\hfil \tcp*{\small{Fit partitions $\CPinputSpace_k$ to $\min \text{Var}(\scoreVal)$}}
$k_i \leftarrow \texttt{DTree}(\CPinput_i)\ \ \forall\{\CPinput_i\}\in\calibset^{CP}$\hfil \tcp*{\small{Assign data to each partition}}
$\hat{q}_k \leftarrow \texttt{SplitCP}(\{\scoreVal_i:k_i=k\};\failureRate),\ \ \forall k$\label{algLine:qhatsplitcp}\hfil \tcp*{\small{Get threshold per leaf node}}
\tcc{Define $\calibDynamics$ as composition of $\approxDynamics$, \texttt{DTree}, \texttt{CAE}. See Sec.~\ref{\detokenize{sec:methodexchange}}.}
\BlankLine
\tcc{(Online) MPC with trajectory optimizer. See Sec.~\ref{\detokenize{sec:methodPlanning}}.}

\While{$(\position,\velocity) \notin$ \texttt{Goal} $\land (\position,\velocity)\in \neg$\texttt{Unsafe}}{
    $\texttt{Unsafe}_t\leftarrow \texttt{UpdateMap}(\position,\projectingFunc{\observation})$ \tcp*{\small{Update planar map}}
    $a^*_{t:t+H-1} \leftarrow \texttt{Plan}(\position,\velocity,\calibDynamics,J,\texttt{Unsafe}_t, H)$\tcp*{\small{Apply $\action^*$ on robot}}
}

\end{algorithm}
}%

\subsection{Observation Processing and Encoding}\label{sec:obsprocessing}
\vspace{-2mm}
Raw camera observations $(\observation^{depth},\observation^{semantics})$ can contain information that is irrelevant for next-state uncertainty quantification.
To facilitate the construction of useful partitions of input space $\CPinputSpace$, we first convert $\observation$ into a robot-frame planar semantic footprint $\observationCone$, shown in Fig. \ref{fig:offlineMethod}.
Let us denote this process by $\projectingFunc{\observation} := \observationCone$.
Using a pinhole camera model and known camera intrinsics, we can define a 3D ray per image pixel, traveling from the camera origin along a fixed robot-frame direction \cite{hartley2003multiple}.
We travel along each ray up to the corresponding $\observation^{depth}$ distance measurement, yielding a 3D point with semantics given by the corresponding $\observation^{semantics}$ pixel.
Given a known camera pose, these 3D points are projected into the ground plane, forming a planar semantic camera footprint region $\observationCone$ ahead of the robot (with angular extent determined by the sensor's FOV and radial extent by the maximum depth range). 
To avoid obstacle splattering \cite{oliveira2015multimodal}, we apply an occlusion-aware rule: obstacles are represented, in the plane, only by their visible boundary (labeled \textsc{occupied}). The region lying behind obstacles' boundaries is kept as \textsc{unknown}.
Finally, we map raw semantic labels into a canonical $\numSemanticClass$-class set $\labelSet = \{\textsc{occupied}, \textsc{unknown}, \textsc{label3}, \ldots, \textsc{label}\numSemanticClass \}$, where the names are purely for interpretability. The footprint is then one-hot encoded into $\observationCone \in \mathbb R^{\numSemanticClass \times H\times W}$, with all classes being treated equally by the following stages.
While the world can be non-flat, this projection implicitly assumes that ground-plane semantics are sufficient for reasoning about short-horizon dynamics transitions.
The resulting semantic footprint is a robot-aligned representation of the nearby environment, yet it may still be too high-dimensional for partition generation. We now consider how to compress $\observationCone$.\newline\indent
We use $\{(\observationCone)_i\}_{i=1}^{\lvert \calibset^{part}\rvert }$ as the training/validation data for a Convolutional Autoencoder (CAE) that learns a low-dimensional latent representation $\latent \in \latentSpace$ of the planar region ahead of the robot.
The CAE predicts pixelwise per-class likelihoods and is trained using cross-entropy reconstruction loss over the semantic classes.
After training, the decoder is discarded, enabling the following compression $\latent = \encoder(\observationCone)$.
Finally, we define our data processing function as
$\processFunc{}: (\position, \velocity,\action,\observation) \mapsto (\velocity,\action,\latent):=(\velocity,\action,\encoder(\projectingFunc{\observation}))$.
\vspace{-5mm}
\subsection{Exchangeability Preservation and
$\CPinputSpace$ Partitioning}\label{sec:methodexchange}
\vspace{-2mm}
Having constructed a lower-dimensional robot-frame representation of factors impacting future transitions, we must now partition the learned latent space $\CPinputSpace$.
For each example in $\calibset^{part}$, we compute its corresponding 
$\CPinput_i = \processFunc{\CPinput_i^{raw}}$, and the nonconformity score $\scoreVal_i := \scoreFunc(\approxDynamics(\CPinput^{raw}_i),\CPoutput_i)$.
Considering the objective of breaking $\CPinputSpace$ up into regions of high and low model uncertainty, we use the LOCART algorithm \cite{locart} and build a Regression Decision Tree (DTree) with inputs $\CPinput_i$ and targets $\scoreVal_i$.
By using the CART splitting process \cite{breiman2017classification},
this tree creates axis-aligned splits to minimize the resulting $\scoreVal$ variance.
Effectively, it partitions $\CPinputSpace$ into regions of approximately constant $\scoreVal$.
Once trained, a forward pass on the DTree then maps a query $\CPinput_i$ to a leaf node, and consequently a given input-space partition $\CPinputSpace_k$.
Thus, we now have a pipeline to transform raw data $\CPinput^{raw}$ into a learned representation $\CPinput$ and assign it to a learned partition $\CPinputSpace_k$.

\looseness-1 We now show this pipeline preserves exchangeability.
From Assumption \ref{ass:exchangeable}, $\calibset:=\{\CPinput_i^{raw},\CPoutput_i\}_{i=1}^\numcal$ is exchangeable with $(\CPinputTest^{raw},\CPoutputTest)$.
It follows directly that $\calibset^{CP} \subset \calibset$ is also exchangeable with raw test transitions \cite{angelopoulos2025theoreticalfoundationsconformalprediction}.
However, our algorithm takes in data processed after the camera projection and CAE encoding.
The calibration dataset becomes $\bar{D}_{cal}^{CP} :=\{\processFunc{\CPinput_i^{raw}}, \CPoutput_i\}_{i=1}^{\lvert \calibset^{CP} \vert}$ and, likewise, test-transitions become $(\processFunc{\CPinputTest^{raw}}, \CPoutputTest)$. To prove $\bar{D}_{cal}^{CP}$ is still exchangeable with processed transitions, consider the following relevant property. 
\begin{proposition}[Proposition 4 of \cite{kuchibhotla2020exchangeability}] \label{prop:permutInvar}
    Suppose $(\CPinput^{raw}_1, \CPoutput_1,\ldots, \CPinput^{raw}_{n+1}, \CPoutput_{n+1})$ are exchangeable, and $g$ is a function depending on $(\CPinput^{raw}_1, \CPoutput_1,\ldots, \CPinput^{raw}_{n+1}, \CPoutput_{n+1})$ permutation invariantly, i.e., $g$ does not use the indexing $i$ to calculate its output. Then $(g(\CPinput^{raw}_1, \CPoutput_1),\ldots, g(\CPinput^{raw}_{n+1}, \CPoutput_{n+1}))$ are exchangeable.
\end{proposition}
That is, if we apply a fixed function to the calibration and test input-output pairs, treating all pairs symmetrically, then we can maintain exchangeability.
Practically, since we do not know $\CPoutputTest$, it would simplify the construction of $\CPregion$ if $g$ does not modify the output.
 Like $g$, $\processFunc{}$ does not consider the index of its test-time or $\calibset^{CP}$ inputs when processing them. It follows that:
\begin{corollary}\label{cor:exchangeableProcessing}
     By Prop. \ref{prop:permutInvar}, $\bar{D}_{cal}^{CP}$ and $(\processFunc{\CPinputTest^{raw}}, \CPoutputTest)$ are exchangeable.
\end{corollary}
\looseness-1 Note that this argument does not hold between $\calibset^{part}$ and test-time transitions, since $\processFunc{}$
depends on $\calibset^{part}$ through the CAE training, yet it is a fixed mapping when applied to $(\processFunc{\CPinputTest^{raw}}, \CPoutputTest)$.
Following now the argument of \cite{locart,lucca}, the partitioning function (forward pass of DTree) assigning processed inputs to partitions $\CPinput \mapsto \CPinputSpace_k$ also acts permutation invariantly between $\bar{D}_{cal}^{CP}$ and $(\CPinputTest, \CPoutputTest)$, since neither of these were used in fitting the tree.
Additionally, since $\bar{D}_{cal}^{CP}$ and $(\CPinputTest, \CPoutputTest)$ are exchangeable, then they are also exchangeable conditioned on a specific partition $\CPinputSpace_k$ \cite{locart}, i.e., if the full data is exchangeable, then the subset of the data corresponding to the $k$-th partition is also exchangeable \cite{angelopoulos2025theoreticalfoundationsconformalprediction,locart}.
Thus, we can perform standard SplitCP on each partition, considering only the $\bar{D}_{cal}^{CP}$ that landed on $\CPinputSpace_k$ to determine the $\hat{q}_k\in \mathbb R$ threshold \cite{locart,lucca}.
We have shown how to achieve \textit{finite-sample local coverage guarantees} using a learned CP input-space.
According to \cite{locart}, if the LOCART partitions $\CPinputSpace_k$ become ``well-populated'' and ``sufficiently thin'' as $\numcal \to \infty$, then their algorithm achieves \textit{asymptotic conditional coverage}.
We make no such claims and leave a rigorous analysis of the asymptotic properties of \methodName~for future work.

\vspace{-3mm}
\subsection{Example Application in Trajectory Optimization}
\label{sec:methodPlanning}
\vspace{-3mm}
We assume perfect perception so that observed obstacles are adequately identified and located.
Starting from a fully unknown environment, we build a planar semantic map at each step using the camera footprint $\observationCone$, the known current $\position$, and heading. Fig. 
\ifarxiv
\ref{fig:isaacSensingExampleDumpsters}
\else
7 of \cite{wafr26arxiv}
\fi
shows an example of the map built after one sensing step.
Let \texttt{Unsafe}${}_t$ denote the subset of the observed regions with label \textsc{occupied}.
Consider navigating towards a known goal set, while avoiding \texttt{Unsafe}${}_t$.
This is generally non-trivial given approximate dynamics of unknown fidelity, as plans we believe to be safe might lead to collision.
Given our linear Gaussian dynamics model $\approxDynamics$, a natural choice for $\scoreFunc$ is the Mahalanobis distance between the predictive distribution and the true next-state.
Following \cite{lucca}, we define a per-partition \textit{conformal scaling factor}\footnote{$\chi^2_{\dim(\stateSpace),\failureRate}$ is the $(1-\failureRate)$ quantile of the chi-squared distribution of dimension $\dim(\stateSpace)$.} $\xi_k :=\hat{q}_k / \chi^2_{\dim(\stateSpace),\failureRate} \in \mathbb R$ which is larger for more uncertain partitions and can be computed offline (see Alg.~\ref{alg:ocular}).
Afterwards, at planning time, each query $\CPinput$ can be passed through the fitted decision tree to determine its corresponding $\xi_k$.  
Starting from the uncalibrated predictive distribution
$\tilde{\mathcal{N}}_{t+1}:=(\tilde{\mu}_{t+1},\tilde{\Sigma}_{t+1})=\approxDynamics(\tilde{\Sigma}_t,\action)$, \cite{lucca} showed that by multiplying the uncalibrated covariance by $\xi$, we can make it \textit{provably calibrated}. That is, the $(1-\failureRate)$ prediction region of $\tilde{\mathcal{N}}_{t+1,cal}:=(\tilde{\mu}_{t+1},\xi \tilde{\Sigma}_{t+1})=\calibDynamics(\tilde{\Sigma}_t,\action)$, a hyperellipsoid, will contain the future state at the user-specified probability.
This occurs as the $(1-\failureRate)$ prediction region of $\tilde{\mathcal{N}}_{t+1,cal}$ corresponds to the $\CPregion$ constructed using Eq \eqref{eq:cpRegionConstruction}, when $\scoreFunc$ is the Mahalanobis distance.
Formally:
\begin{theorem}[Theorem 2 of \cite{lucca}] The $(1-\failureRate)$ prediction region of the calibrated multivariate normal
    $\tilde{\mathcal{N}}_{t+1,cal}:=(\tilde{\mu}_{t+1},\tilde{\Sigma}_{t+1,cal})$ contains the next robot state $\state{}_{+1} \sim f(\state,\action)$ with a likelihood greater than or equal to $(1-\failureRate)$.
\end{theorem}
Thus, if the $(1-\failureRate)$ prediction region of $\tilde{\mathcal{N}}_{t+1,cal}$ does not intersect \texttt{Unsafe}${}_t$, and is contained in the previously viewed areas\footnote{There might be obstacles in the unobserved regions.}, then we know the planned action $\action$ is probabilistically safe, since $\mathbb P\{\CPoutputTest \in \CPregion \}\ge(1-\failureRate) \land (\CPregion \cap \texttt{Unsafe}_t = \emptyset) \Rightarrow \mathbb P\{\CPoutputTest \notin \texttt{Unsafe}_t \} \ge (1-\failureRate)$.
 \methodName~ then achieves probabilistic one-step safety up to a user-defined acceptable failure-rate $\failureRate \in (0,1)$ by using the same calibrated dynamics $\calibDynamics$.
 After the first planning step, we no longer have access to raw observations $\observation$ and instead
approximate $\observationCone{}_{+h}$ by perspective-projecting the planar semantic map built during execution using the estimated future pose $\tilde{p}_{t+h}$ and heading.
 We use $(\tilde{p}_{t+h},\tilde{v}_{t+h}):=\mathbb E[\tilde{\mathcal N}_{t+h}]$, with velocity providing the direction of motion.
 As in the first step, we concatenate the estimated velocity and the commanded actions with $\observationCone{}_{+h}$ to construct $\CPinput_{t+h}^{raw}$.
 This may then be fed to the trained CAE and DTree to determine its corresponding scaling factor $\xi_k$.
 Our multistep planning heuristic then consists of chaining uncalibrated predictive Gaussian propagations through $\approxDynamics$ with uncertainty calibrations by $\xi_k$.
 The $\tilde{\mathcal N}_{\tau,cal}$ at each step serves as the input for $\approxDynamics$ in the following step $\tau+1$.
 While empirically useful, the map-generated $\observationCone$ is not necessarily exchangeable with the calibration data or the one-step $\observationCone$.
 Hence, we make no theoretical claims about multi-step safety.
 We can then solve the following receding-horizon MPC problem recursively, taking one probabilistically safe action after each solution:
\vspace{-3mm}
{
\setlength{\belowdisplayskip}{3pt}
\setlength{\belowdisplayshortskip}{3pt}
\begin{subequations}
\setlength{\jot}{0pt} 
\begin{alignat}{3}
&\!\min_{(u_{t},\ldots,u_{t+H-1})}        &\qquad& J(s_{t+1:t+H}, u_{t:t+H-1}, \texttt{Goal})\label{eq:optCost}\\[-1mm]
&\text{subject to} & & \CPregion(\CPinput_\tau) \cap \texttt{Unsafe}_t =\emptyset,\quad & \forall\tau \in \{t,\ldots, t+H-1\},\label{eq:optSafety}\vspace{-7mm}
\end{alignat}
\end{subequations}}where \texttt{Goal} is a known region, and $\CPregion$ is the conformal region generated at planning index $\tau$ by our heuristic multi-step uncertainty propagation.
\vspace{-5mm}
\section{Experiments}\label{sec:experiments}
\vspace{-3mm}
\looseness-1 Our experiments aim to validate the one-step local coverage guarantees of \methodName~and to demonstrate its applicability for probabilistically safe planning under both aleatoric disturbances and model mismatch.
We test our method on a two-dimensional double-integrator. 
Experiments are conducted on a planar environment with a simplified sensor $(\text{Sec.}~\ref{sec:expPlanar})$ and in Isaac Sim using a floating-camera attached to the double-integrator $(\text{Sec.}~\ref{sec:expIsaac})$, to demonstrate our method's ability to handle more realistic perception information.
The system state can be decomposed into 2D positions $\position:=(p_{x,t},p_{y,t})$ and velocities $\velocity:=(v_{x,t},v_{y,t})$. The actions are bounded accelerations $\action:=(a_{x,t},a_{y,t})\in [-0.9,0.9]^2$.
Under nominal conditions -- shown as green in Fig. \ref{fig:runsMapU} and \ref{fig:runsIcyMainIsaac} -- the true system $\trueDynamics$ evolves according to the standard double-integrator equations -- see Eq 
\ifarxiv
\eqref{eq:doubleIntegrator} of App. \ref{app:dynamics}.
\else
(B.1) of  App. B of \cite{wafr26arxiv}.
\fi
Stochasticity arises from random external disturbances.
However, in some regions of the environment  -- shown as white in Fig. \ref{fig:runsMapU} and \ref{fig:runsIcyMainIsaac} -- the true system $\trueDynamics$ evolves according to the shifted dynamics in Eq
\ifarxiv
\eqref{eq:shiftedIntegratorDynamics},
\else
(B.2)
\fi
where the multiplicative $\times 1.3$ term amplifies both control-induced displacements and uncontrolled $(\action=0)$ motion, leading to a faster accumulation of momentum. This represents slippery/lower-friction surfaces.
\looseness-1 The shift in dynamics from 
\ifarxiv
Eq.~\eqref{eq:doubleIntegrator}
\else
Eq.~(B.1)
\fi
to 
\ifarxiv
Eq.~\eqref{eq:shiftedIntegratorDynamics}
\else
Eq.~(B.2)
\fi
occurs when changing regions, making $\trueDynamics$ hybrid.
The linear Gaussian dynamics model $\approxDynamics$ follows Eq
\ifarxiv
\eqref{eq:doubleIntegrator},
\else
(B.1),
\fi
being in-distribution (ID) in the green regions and OOD in the white regions.
Hence, the double-integrator requires accurate multistep uncertainty estimation, as momentum buildup coupled with uncertainty underestimation can lead to the robot entering regions of inevitable collision (e.g., traveling too fast towards a wall) \cite{lavalle2006planning}. 
We use $\failureRate=0.10$.

\looseness-1
We compare \methodName~with three competitive baselines. No Calibration (\textit{NoCP}) directly uses the $(1-\failureRate)$ prediction region from $\approxDynamics$.
\textit{SplitCP} (Sec.~\ref{sec:SplitCP}) uses $\calibset$ from the tested environment to construct a single scaling factor $\xi$.
\textit{LUCCa} \cite{lucca} also uses $\calibset$ from the tested environment but builds a LocalCP partitioning on $X:=(\position,\velocity,\action)$ with a DTree, leading to per-leaf-node factors $\xi_k$.
Ours is the only CP method using \textit{no data from the tested environment}, while \textit{SplitCP} and \textit{LUCCa} have access to test-environment data.
In App.
\ifarxiv
\ref{app:additionalResults},
\else
D of \cite{wafr26arxiv},
\fi
we additionally compare \methodName~with two ablations: 1) OCULAR using only data from the tested environment, as do \textit{SplitCP} and \textit{LUCCa}; 2) OCULAR without a learned lower-dimensional observation embedding, i.e., the DTree input becomes $X_i=(v_t,a_t)$. The former studies the impact of dataset size and source (which environment $\calibset$ comes from), and the latter the impact of perception information. 
\vspace{-4mm}
\subsection{Planar Robot: Slippery Corridors}
\label{sec:expPlanar}
\vspace{-2mm}
\looseness-1 Both $\trueDynamics$ and $\approxDynamics$ use $\Delta t= 0.05$ s and process noise $Q=0.1BB^\top+10^{-6}I$. Observations are obtained from a simplified ground-truth sensor measuring a conic section of the environment along the robot's direction of motion. 
We consider a FOV of 120 deg and maximum depth range of 0.65 m, producing a discretized polar image with a resolution of $240\times 80$ (angle$\times$depth).
Each pixel corresponds to a fixed robot-frame position in polar coordinates, with occlusions handled via the masking rule of Sec. \ref{sec:obsprocessing}. We evaluate on 4 maps shown in Fig.
\ifarxiv
\ref{fig:planarEnvInference}
\else
5 of \cite{wafr26arxiv}
\fi
(S, U, L, and H).

\looseness-1 $\textsc{lin}(a,b,N)$ denotes a linearly-spaced sequence of $N$ real values between $a$ and $b$.
A calibration dataset $\calibset$ was collected per environment, by enumerating the grid
$(p_x,p_y) \in \textsc{lin}(map_x^{\min},map_x^{\max},16)\times \textsc{lin}(map_y^{\min},map_y^{\max},16)$,
$(v_x,v_y) \in \textsc{lin}(-1.8,1.8,8)^2$,
$(a_{x,t},a_{y,t}) \in \textsc{lin}(-0.9,0.9,4)^2$, rolling out one-step transitions using $\trueDynamics$, and collecting observations $\observation$.
Transitions that collided were discarded, hence environments with smaller free-space areas have smaller $\lvert \calibset \rvert$: 90.9k for H; 71.3k for L; 79.9k for S; 146.7k for U.
Since LUCCa and SplitCP use data from the tested environment, these transitions constitute their respective calibration sets.
\methodName~uses data from all but the tested environment (e.g., for H, our method uses data from S, L, U only), leading to: $\lvert \calibset\rvert=228.4$k for H; $\lvert \calibset\rvert=241.6$k for L; $\lvert \calibset\rvert=229.2$k for S; $\lvert \calibset\rvert=176.4$k for U.
While our calibration dataset is larger, it only contains data from \textit{different but visually-similar} environments. Both LUCCa and \methodName~use $70\%$ of $\calibset$ for $\calibset^{part}$ (the rest for $\calibset^{CP}$) and for the DTree a max depth $13$ and min samples leaf $300$. \methodName's symmetric CAE uses a three-layer encoder with channel widths $\{8, 16, 32\}$, $4\times4$ kernels, stride $2$, and padding $1$, producing a six-dimensional $\latent$.\newline\indent
\textbf{Numerical Coverage Validation.} To validate the key coverage claim, we evaluate the four methods on thousands of true dynamics transitions generated analogously to $\calibset$ by enumerating a grid with the same state–action ranges but coarser discretization.
For each test state–action pair, we propagate $10{,}000$ Monte Carlo (MC) particles under $\trueDynamics$ to estimate the \emph{empirical coverage}, defined as the fraction of particles lying inside a method’s prediction region.
This represents the likelihood of said region containing the true next unknown state $\state{}_{+1}$. By averaging empirical coverage over test cases, we get a numerical estimate of \textit{marginal coverage}, the key guarantee of Eq \eqref{eq:marginalCoverage}.
We report in Table \ref{tab:test2case2NOabblation} the marginal coverage \emph{conditioned on region type}, computing separate averages over test cases lying in ID and OOD regions.
Methods guaranteeing global marginal coverage (e.g., SplitCP) may still exhibit poor coverage when restricted to specific subsets of the state space, while local coverage methods are expected to perform well even in under-represented or OOD subsets.
Additionally, we compute for each test-case, using an ``oracle'' dynamics model, the smallest possible Gaussian prediction region achieving $90\%$ coverage that is centered at the uncalibrated mean $\tilde{\mu}_{\tau}$ of $\approxDynamics$ and has covariance obtained by isotropically scaling the uncalibrated covariance. This represents an unachievable ideal, indicating how close to ``optimal'' the different methods are under our problem constraints. In Table \ref{tab:test2case2NOabblation} we report the median relative volume between each method and this oracle, serving as a measure of prediction region \textit{volume-efficiency}.
\begin{table*}[!t]
\captionsetup{skip=4pt}
\centering
\scriptsize
\setlength{\tabcolsep}{1.2pt}
\renewcommand{\arraystretch}{1.12}
\caption{Test-case results across four planar environment maps.}
\label{tab:test2case2NOabblation}

\begin{tabular}{c | l | c |
  cc @{\hspace{3pt}}!{\vrule width 0.3pt}@{\hspace{3pt}}
  cc @{\hspace{3pt}}!{\vrule width 0.3pt}@{\hspace{3pt}}
  cc @{\hspace{3pt}}!{\vrule width 0.3pt}@{\hspace{3pt}}
  cc}
\toprule
\multirow{2}{*}{Metric}
& \multicolumn{1}{c|}{\multirow{2}{*}{Method}}
& \multirow{2}{*}{\shortstack{Tested map\\\textbf{not} in $\calibset$?}}
& \multicolumn{2}{c@{\hspace{3pt}}!{\vrule width 0.3pt}@{\hspace{3pt}}}{Map S}
& \multicolumn{2}{c@{\hspace{3pt}}!{\vrule width 0.3pt}@{\hspace{3pt}}}{Map L}
& \multicolumn{2}{c@{\hspace{3pt}}!{\vrule width 0.3pt}@{\hspace{3pt}}}{Map H}
& \multicolumn{2}{c}{Map U} \\
& & & ID & OOD & ID & OOD & ID & OOD & ID & OOD \\
\midrule

\multirow{4}{*}{\shortstack{Marginal\\Coverage (\%)}}
& NoCP & N/A & \ccell{red!15}{89.9} & \ccell{red!15}{6.4} & \ccell{green!15}{90.0} & \ccell{red!15}{6.4} & \ccell{green!15}{90.0} & \ccell{red!15}{6.4} & \ccell{red!15}{89.9} & \ccell{red!15}{6.4} \\
& SplitCP & \myxmark & \ccell{green!15}{100.0} & \ccell{red!15}{69.7} & \ccell{green!15}{100.0} & \ccell{red!15}{69.1} & \ccell{green!15}{100.0} & \ccell{red!15}{61.8} & \ccell{green!15}{100.0} & \ccell{red!15}{71.6} \\
& LUCCa \cite{lucca} & \myxmark & \ccell{green!15}{91.4} & \ccell{green!15}{93.4} & \ccell{green!15}{91.1} & \ccell{green!15}{93.8} & \ccell{green!15}{91.0} & \ccell{green!15}{93.9} & \ccell{green!15}{90.6} & \ccell{green!15}{93.8} \\
& \textbf{OCULAR (ours)} & \mycheck & \ccell{green!15}{90.6} & \ccell{green!15}{93.4} & \ccell{green!15}{90.8} & \ccell{green!15}{93.7} & \ccell{green!15}{91.2} & \ccell{green!15}{91.0} & \ccell{green!15}{90.5} & \ccell{green!15}{93.8} \\
\midrule

\multirow{4}{*}{\shortstack{Median\\$\CPregion$ volume\\(wrt oracle) $\downarrow$}}
& NoCP & N/A & 0.99 & 0.02 & {1.00} & 0.02 & {1.00} & 0.02 & 0.99 & 0.02 \\
& SplitCP & \myxmark & 50.26 & 0.84 & 49.66 & 0.83 & 40.50 & 0.67 & 53.28 & 0.89 \\
& LUCCa \cite{lucca} & \myxmark & 1.11 & 1.14 & 1.06 & {1.14} & 1.09 & {1.11} & 1.06 & {1.14} \\
& \textbf{OCULAR (ours)} & \mycheck & {1.00} & {1.11} & 1.05 & {1.15} & {1.00} & 1.13 & {1.05} & {1.15} \\
\bottomrule
\end{tabular}
\vspace{2pt}
\begin{minipage}{\linewidth}
\scriptsize
\ifarxiv
\colorbox{red!15}{red}: coverage below $(1-\failureRate)=0.9$.
\else
Coverage $<90\%$: undercoverage.
\fi\ $\CPregion$ volume reported relative to oracle using the minimum scaling $\xi$ needed to achieve $0.9$ coverage per transition.
Test transition \#: S 4,096; L 2,816; H 4,096; U 6,656.
\end{minipage}
\vspace{-7mm}
\end{table*}

\newline\indent Unsurprisingly, \textit{NoCP} severely underestimated the real uncertainty of OOD regions. \textit{SplitCP}'s lack of adaptivity (single $\xi$) made it overconservative in ID and over-optimistic when OOD. 
Both LUCCa and \methodName~performed comparatively well, providing sufficient coverage in both ID and OOD scenarios and constructing prediction regions that were only $10-16\%$ larger than the ``oracle'' prediction-region when OOD.
While methods with coverage above the user-specified threshold of $0.9$ are equally calibrated, too high coverage (e.g., \textit{SplitCP} when ID) indicates an over-conservative calibration procedure, since $\CPregion_1 \subseteq \CPregion_2 \Rightarrow \text{Coverage}_1 \le \text{Coverage}_2$. 
Given the practical objective of constructing a sufficiently calibrated region that is as volume efficient as possible, \methodName~appears to achieve this equally well to methods that have environment-specific data, validating the quality of our learned representations.

\looseness-1 \textbf{Safe Planning with Calibrated Uncertainty Estimates.} Despite only guaranteeing one-step safety, we perform MPC trajectory optimization experiments to validate the usefulness of the heuristic multi-step uncertainty propagation proposed in Sec.~\ref{sec:methodPlanning}.
At each step, we build a map by majority voting using $\observationCone$ on a discretized grid map with a resolution of $0.01$ m. We ignore \textsc{unknown} sensor points when updating the map seen up to time $t$.
We used MPPI \cite{williams2017information} as the sampling-based trajectory optimizer (see App.
\ifarxiv
\ref{app:plannerdetails}
\else
C of \cite{wafr26arxiv}
\fi
for hyperparameters).
The objective function, Eq
\ifarxiv
\eqref{eq:mppiCost},
\else
(C.1) of \cite{wafr26arxiv},
\fi
balances task progress with collision avoidance, while steering plans towards $\CPinputSpace_k$ with lower uncertainty.
Fig. \ref{fig:runsMapU} shows the trajectories on map U and App. 
\ifarxiv
\ref{app:additionalResultsPlanar}
\else
D.1 of \cite{wafr26arxiv}
\fi
the trajectories on all environments.  The project website includes execution videos. Table \ref{tab:planning_all_maps_compact} reports the success rate of each method (percentage of trials where all subgoals are reached without any collisions) and the average number of steps taken to complete each trial.
{
\begin{figure}[!b]
\vspace{-8mm}
\includegraphics[width=\textwidth]{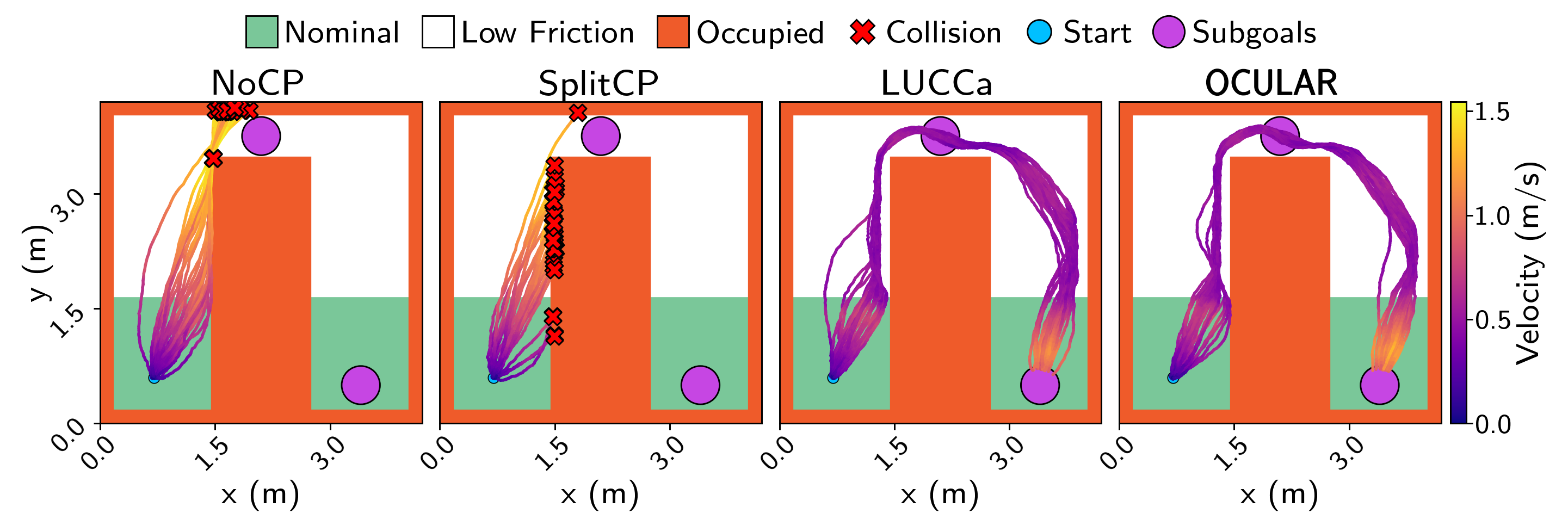}\vspace{-3.5mm}
\caption{\looseness-1 Rollouts in map U of the planar robot (see
App.
\ifarxiv
\ref{app:additionalResultsPlanar}
\else
D.1 of \cite{wafr26arxiv}
\fi
for all  maps). \textit{NoCP} gains too much momentum in the low-friction region, leading to collisions. \textit{SplitCP}'s single scaling factor makes all its sampled plans highly conservative and in collision, resulting in goal-chasing behavior.
\textit{LUCCa} and \methodName~have comparable performance, slowing down in high-uncertainty regions, and reaching all subgoals safely. Yet, \textit{LUCCa} requires data specific to each tested map, while our method produces safe plans \textit{without any data from the executed environment.}
} \label{fig:runsMapU}\vspace{-6mm}
\end{figure}
}
\begin{table*}[!t]
\captionsetup{skip=4pt}
\centering
\scriptsize
\setlength{\tabcolsep}{1.3pt}
\renewcommand{\arraystretch}{1.22} 
\caption{Planning results across four planar environment maps ($30$ runs each).}
\label{tab:planning_all_maps_compact}
\begin{tabular}{c||c||c|c|c|c||c@{\hspace{3.5pt}}|c|c|c}
\toprule
\multirow{2}{*}{\centering Method}
& \multirow{2}{*}{\shortstack{Tested map\\\textbf{not} in $\calibset$?}}
& \multicolumn{4}{c||}{Success (\%) $\uparrow$}
& \multicolumn{4}{c}{Steps to completion (mean$\pm$std) $\downarrow$} \\
& & S & L & H & U & S & L & H & U \\
\midrule

\multicolumn{1}{l||}{NoCP}
& N/A
& 33.3 & 0 & 6.7 & 0
& \makebox[2.65em][r]{214.8}$\pm$\makebox[1.75em][l]{39.3} & -- & \makebox[2.65em][r]{155.5}$\pm$\makebox[1.75em][l]{0.7} & -- \\
\multicolumn{1}{l||}{SplitCP}
& \myxmark
& 0 & 0 & 0 & 0
& -- & -- & -- & -- \\
\multicolumn{1}{l||}{LUCCa \cite{lucca}}
& \myxmark
& 100 & 100 & 100 & 100
& \makebox[2.65em][r]{{171.1}}$\pm$\makebox[1.75em][l]{{12.1}}& \makebox[2.65em][r]{{211.2}}$\pm$\makebox[1.75em][l]{{6.5}} & \makebox[2.65em][r]{{203.2}}$\pm$\makebox[1.75em][l]{{6.6}} & \makebox[2.65em][r]{{283.3}}$\pm$\makebox[1.75em][l]{{8.1}} \\
\multicolumn{1}{l||}{\textbf{OCULAR}}
& \mycheck
& {100} & {100} & {100} & {100}
& \makebox[2.65em][r]{{177.6}}$\pm$\makebox[1.75em][l]{{7.0}} & \makebox[2.65em][r]{{213.6}}$\pm$\makebox[1.75em][l]{{7.5}} & \makebox[2.65em][r]{{199.7}}$\pm$\makebox[1.75em][l]{{7.7}} & \makebox[2.65em][r]{{278.1}}$\pm$\makebox[1.75em][l]{{7.6}} \\
\bottomrule
\end{tabular}
\begin{minipage}{\linewidth}
\scriptsize
Success $=$ reaching all subgoals without collisions. 
\end{minipage}
\vspace{-10mm}
\end{table*}

\newline\indent \textit{NoCP} gains too much momentum when OOD, often colliding due to being over-optimistic when travelling at high speeds, not satisfying the user-provided safety requirement. \textit{SplitCP}'s significant conservativeness makes all its plans end in collision, leading to poor obstacle avoidance.
\textit{LUCCa} and \methodName~appear to keep uncertainty low when OOD by slowing down and achieve efficient collision-free trajectories.
These results indicate that our approach can generalize to new unseen environments and can achieve adequate planning performance and safety by leveraging perception information obtained elsewhere.

\vspace{-5mm}
\subsection{Isaac Sim: Snow-Covered T-Junction}\label{sec:expIsaac}
\begin{figure}[!b]
\vspace{-6mm}
\includegraphics{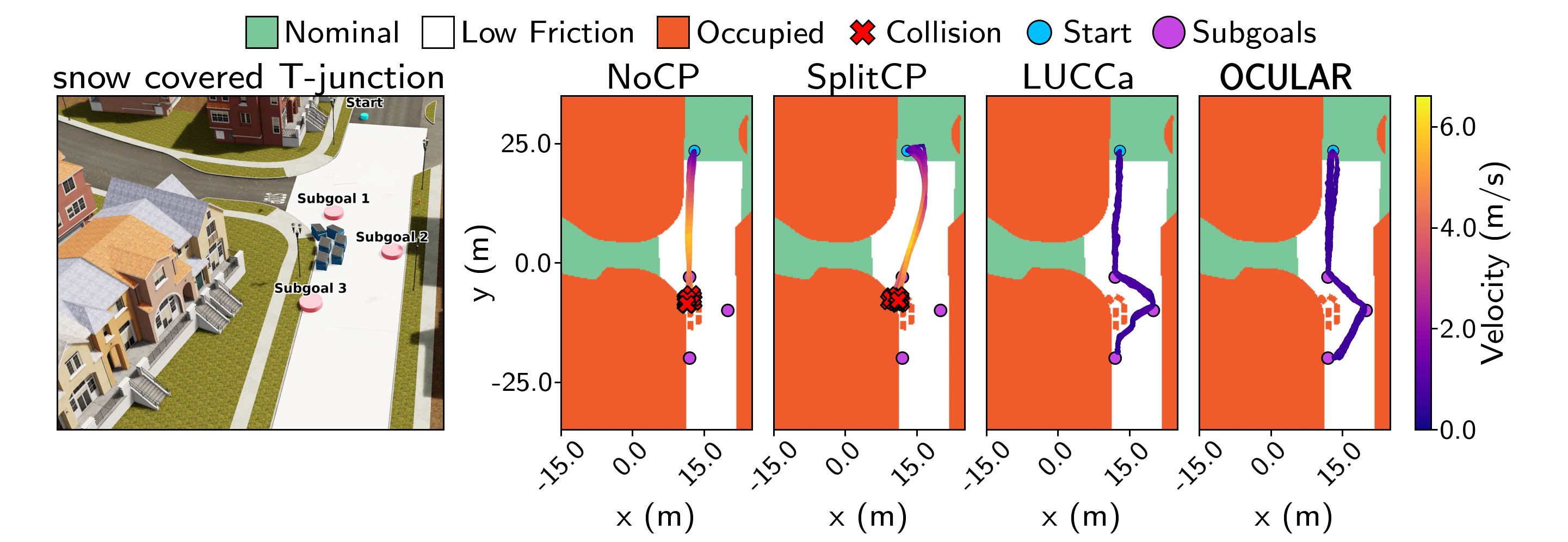}\vspace{-4mm}
\caption{\looseness-1 Rollouts in icyMain of  Isaac Sim experiment (see App.
\ifarxiv
\ref{app:additionalResultIsaac}
\else
D.2 of \cite{wafr26arxiv}
\fi
for other maps). \textit{NoCP} and \textit{SplitCP} gain too much momentum over ice and collide.
\textit{LUCCa} and \methodName~both slow down when OOD, keeping next-state uncertainty manageable and reaching all subgoals without collisions. While \textit{LUCCa} requires icyMain-specific data, our method plans \textit{without any data from icyMain.}
}\label{fig:runsIcyMainIsaac}\vspace{-12mm}
\end{figure}

\vspace{-3mm}
\looseness-1 Observations come from ground-truth depth and semantic segmentation cameras
attached to the double-integrator.
Both were captured at $240\times240$ resolution, with max depth range of 12 m. Fig.
\ifarxiv
\ref{fig:isaacSensingExampleDumpsters}
\else
7 of \cite{wafr26arxiv}
\fi
shows an example observation $\observation$, planar cone representation $\observationCone$, and map update after one step.
We consider 3 environments named icySide, icyMain, and icyMiddle (cf. Fig.
\ifarxiv
\ref{fig:isaacSimMaps}).
\else
6 of \cite{wafr26arxiv}).
\fi
Due to the larger map size, $\trueDynamics$ and $\approxDynamics$ use $\Delta t=0.2$\,s and disturbance covariance $Q=0.1BB^\top$.
The calibration dataset $\calibset$ was again generated by spanning a collision-free Cartesian grid and rolling out one-step transitions under $\trueDynamics$. \methodName~used the same hyperparameters as in Sec. \ref{sec:expPlanar}, and we re-tuned the baselines.\newline\indent
\begin{table*}[!b]
\vspace{-10mm}
\captionsetup{skip=4pt}
\centering
\scriptsize
\setlength{\tabcolsep}{2pt}
\renewcommand{\arraystretch}{1.12}
\caption{Test-case results across three Isaac Sim roads.}
\label{tab:isaacTestCases}
\begin{tabular}{c | l | c |
  cc @{\hspace{3pt}}!{\vrule width 0.3pt}@{\hspace{3pt}}
  cc @{\hspace{3pt}}!{\vrule width 0.3pt}@{\hspace{3pt}}
  cc}
\toprule
\multirow{2}{*}{Metric}
& \multirow{2}{*}{\makebox[1.4cm][c]{Method}}
& \multirow{2}{*}{\shortstack{Tested map\\\textbf{not} in $\calibset$?}}
& \multicolumn{2}{c@{\hspace{3pt}}!{\vrule width 0.3pt}@{\hspace{3pt}}}{icyMain}
& \multicolumn{2}{c@{\hspace{3pt}}!{\vrule width 0.3pt}@{\hspace{3pt}}}{icyMiddle}
& \multicolumn{2}{c}{icySide} \\
& & & ID & OOD & ID & OOD & ID & OOD \\
\midrule

\multirow{4}{*}{\shortstack{Marginal\\Coverage (\%)}}
& NoCP
& N/A
& \ccell{green!15}90.0 & \ccell{red!15}56.7
& \ccell{green!15}90.0 & \ccell{red!15}56.7
& \ccell{green!15}90.0 & \ccell{red!15}56.7 \\

& SplitCP
& \myxmark
& \ccell{green!15}99.8 & \ccell{green!15}93.0
& \ccell{green!15}99.1 & \ccell{red!15}85.9
& \ccell{green!15}99.5 & \ccell{red!15}89.6 \\

& LUCCa \cite{lucca}
& \myxmark
& \ccell{green!15}90.1 & \ccell{green!15}91.4
& \ccell{green!15}90.1 & \ccell{green!15}90.9
& \ccell{green!15}91.1 & \ccell{green!15}91.5 \\

& \textbf{OCULAR (ours)}
& \mycheck
& \ccell{green!15}90.4 & \ccell{green!15}91.0
& \ccell{green!15}91.1 & \ccell{green!15}90.6
& \ccell{green!15}91.5 & \ccell{green!15}90.1 \\
\midrule

\multirow{4}{*}{\shortstack{Median\\$\CPregion$ volume\\(wrt oracle) $\downarrow$}}
& NoCP
& N/A
& 1.00 & 0.28
& 1.00 & 0.28
& 1.00 & 0.28 \\

& SplitCP
& \myxmark
& 4.66 & 1.29
& 3.07 & 0.85
& 3.73 & 1.03 \\

& LUCCa \cite{lucca}
& \myxmark
& {1.02} & {1.10}
& {1.02} & 1.13
& 1.08 & 1.13 \\

& \textbf{OCULAR (ours)}
& \mycheck
& {1.02} & 1.15
& 1.06 & {1.06}
& {1.03} & {1.02} \\
\bottomrule
\end{tabular}

\begin{minipage}{\linewidth}
\scriptsize
\ifarxiv
\colorbox{red!15}{red}: coverage below $(1-\failureRate)=0.9$.
\else
Coverage $<90\%$: undercoverage.
\fi\ Volume reported as ratio relative to an oracle using the minimum $\xi$ to achieve $90\%$ coverage.
Test transition \#: icyMain 4,464; icyMiddle 4,464; icySide 4,464.
\end{minipage}
\end{table*}
\textbf{Numerical Coverage Validation.}
Test cases are generated again from a separate grid, now spanning narrower and coarser position and velocity ranges to obtain enough collision-free trajectories that are sufficiently far from the environment boundaries (observations beyond boundaries were deemed unknown).
As in Sec. \ref{sec:expPlanar}, we propagate $10k$ MC particles per test case to estimate marginal coverage \emph{conditioned on region type}.
Table \ref{tab:isaacTestCases} shows the performance of different algorithms across the three environments.
As with the planar robot, \textit{NoCP} drastically undercovers the OOD setting. \textit{SplitCP}, possibly due to a different data imbalance, is sometimes over-conservative (i.e., icyMain). Yet, \textit{SplitCP} lacks \textit{adaptivity} and can still undercover when OOD. Both \textit{LUCCa} and \methodName~achieve the user-set coverage, being clearly more \textit{volume-efficient} than \textit{SplitCP}.
Ultimately, the results indicate that \methodName, which is not given any data in the tested environment, can achieve local coverage with 
comparable or greater volume efficiency than baselines requiring environment-specific data. This suggests that our approach can still generalize when the perception information is \textit{higher-dimensional and more realistic}.
We additionally performed probabilistically safe trajectory optimization experiments in Isaac Sim, using the proposed multi-step uncertainty propagation of Sec.~\ref{sec:methodPlanning}. The numerical and qualitative results (App.
\ifarxiv
\ref{app:additionalResultIsaac})
\else
D.2 of \cite{wafr26arxiv})
\fi suggest that \methodName~can produce plans that are comparably safe and efficient to those generated by methods with access to data from the test environment.
Fig. \ref{fig:runsIcyMainIsaac} shows the rollouts observed in icyMain.

\vspace{-6mm}
\subsection{Limitations}\vspace{-3mm}
\looseness-1 Despite showing that \methodName~performs equally well in \textit{unseen} environments as methods with access to environment-specific data, our approach is sensitive to the hyperparameters of the CAE and DTree (as is LUCCa \cite{lucca}).
For some hyperparameter choices, \methodName~produces inadequate partitions that can lead to undercoverage. 
Additionally, our latent-representation learning is detached from any uncertainty-prediction utility, i.e., we train the CAE with a pure reconstruction loss. It might be helpful to instead learn latent representations that can facilitate downstream partitioning into regions of high and low uncertainty, though this would require differentiating through the decision tree, which is non-trivial.
\vspace{-6mm}
\section{Conclusion}
\vspace{-4mm}
\looseness-1 We present a method for local conformal prediction from perception information, that accounts for both aleatoric and epistemic uncertainty without having access to any data in the test environment (while existing approaches require environment-specific data).
Our experiments, on a planar robot and in Isaac Sim, suggest that \methodName~can construct prediction regions that are sufficiently calibrated, both ID and OOD, with comparable volume-efficiency to baselines.
Further, we demonstrate how \methodName~might be used in a downstream probabilistic planning task, maintaining safety despite significant epistemic uncertainty.

\begin{credits}
\vspace{-2mm}
\subsubsection{\ackname} This work was supported in part by the Office of Naval Research Grant N00014-24-1-2036 and NSF grants IIS-2113401 and IIS-2220876.
\vspace{-2mm}
\subsubsection{\discintname}
The authors have no competing interests to declare.
\end{credits}

\bibliographystyle{splncs04}
\bibliography{WAFR26-SensorCP}

\ifarxiv
\clearpage
\appendix

\renewcommand{\theequation}{\thesection.\arabic{equation}}
\makeatletter
\@addtoreset{equation}{section}
\makeatother

\section{Additional Problem Motivation}\label{app:extramotivation}
\vspace{-3mm}
In safety-critical scenarios, and for tasks requiring reasoning about uncertainty, user-set confidence levels $(1-\failureRate)\in (0,1)$ are often used to construct \textit{prediction regions}\footnote{For Gaussian $\approxDynamics$, prediction regions take the shape of hyperellipsoids.} containing $(1-\failureRate)\%$ of the predictive probability mass.
For example, one might plan so that the $90\%$ prediction regions do not intersect any known obstacles or plan so that the region's size is reduced, to incentivize safety and to minimize uncertainty, respectively.
However, prediction regions generated from $\approxDynamics$ are generally \textit{uncalibrated}; e.g. a region containing $90\%$ of the predictive probability mass may not contain $90\%$ of the true future states, which evolve according to $\trueDynamics$, not $\approxDynamics$.
This lack of calibration can produce catastrophic consequences.
Overconfident models may lead to execution-time collisions despite apparently safe plans, while underconfident models may be overly conservative and fail to make task progress.
Both cases are undesirable.
Approximate models alone are hence insufficient for safety-critical tasks.
\vspace{-3mm}
\section{Dynamical System Equations}\label{app:dynamics}
\vspace{-2mm}
The standard double-integrator dynamics with Heun discretization \cite{cosner2023robust,lucca} are
\begin{equation}\label{eq:doubleIntegrator}
\begin{bmatrix}
\position{}_{+1}\\
\velocity{}_{+1}
\end{bmatrix}
=
\underbrace{
\begin{bmatrix}
I_{2\times 2}
\hspace{4pt} & \hspace{4pt}
\Delta t\, I_{2\times 2}
\\[4pt]
0_{2\times 2}
\hspace{4pt} & \hspace{4pt}
I_{2\times 2}
\end{bmatrix}}_{A}
\begin{bmatrix}
\position\\
\velocity
\end{bmatrix}
+
\underbrace{
\begin{bmatrix}
\frac{\Delta t^2}{2}\, I_{2\times 2}
\\[4pt]
\Delta t\, I_{2\times 2}
\end{bmatrix}}_{B}
\,\action
+
\disturbance,
\quad
\disturbance \sim \mathcal{N}\!\left(0,\; Q\right),
\vspace{-3mm}
\end{equation}
where $\disturbance$ is independently sampled at each time-step, introducing aleatoric perturbations.
To represent slippery/lower-friction surfaces, the true system evolves according to the following dynamics in the white regions of the planar and Isaac environments:
\begin{equation}\label{eq:shiftedIntegratorDynamics}
\begin{bmatrix}
\position{}_{+1}\\
\velocity{}_{+1}
\end{bmatrix}
=
\underbrace{
\begin{bmatrix}
I_{2\times 2}
\hspace{4pt} & \hspace{4pt}
\mismatch\,\Delta t\, I_{2\times 2}
\\[4pt]
0_{2\times 2}
\hspace{4pt} & \hspace{4pt}
I_{2\times 2}
\end{bmatrix}}_{A_{\text{shift}}}
\begin{bmatrix}
\position\\
\velocity
\end{bmatrix}
+
\underbrace{
\begin{bmatrix}
\mismatch\,\frac{\Delta t^2}{2}\, I_{2\times 2}
\\[4pt]
\mismatch\,\Delta t\, I_{2\times 2}
\end{bmatrix}}_{B_{\text{shift}}}
\,\action
+
\disturbance .
\vspace{-2mm}
\end{equation}
\vspace{-5mm}
\section{Planning Implementation Details}\label{app:plannerdetails}
\vspace{-2mm}
We used Model Predictive Path Integral Control (MPPI) as our trajectory optimizer for all experiments and methods.
The objective function used for probabilistically safe plan generation was 
\begin{equation}\label{eq:mppiCost}\vspace{-1mm}
    J:= c_{dist}^{term}d_{E}(\tilde{\mu}_{t+H}) + \sum_{\tau =t+1}^{t+H-1}c_{dist}^{run}d_E(\tilde{\mu}_\tau) + c_{trace}^{run}\text{tr}(\tilde{\mathcal N}_{\tau,cal})+c_{col}^{run}\indicator_{\{\CPregion(\CPinput_\tau) \cap \texttt{Unsafe}_t \ne \emptyset\}}, \
\end{equation}
where $d_E$ is the Euclidean distance between the predictive Gaussian's mean and the \texttt{Goal} region, $\text{tr}(\cdot)$ the trace of the calibrated Gaussian, and the indicator function determines when the $(1-\failureRate)$ prediction region of the calibrated Gaussian intersects with the obstacles observed at time $t$.
The trace term is intended to steer plans towards regions of input space $\CPinputSpace_k$ with lower dynamics uncertainty.
For the planar system experiments of Sec. \ref{sec:expPlanar}, we used $c_{dist}^{run}=0.8$,
$c_{dist}^{term}=1.0$,
$c_{trace}^{run}=1.5$, $c_{col}^{run} =1,000$, with an MPPI temperature $\lambda=0.25$, $2048$ sampled action-trajectories per step, control noise $I_{2\times 2}$, and horizon of $H=9$. Horizons that are too short can lead to myopic behavior and long horizons may make all trajectories infeasible (due to the successive scaling performed in our heuristic propagation). 
For the Isaac Sim trajectory optimization experiments, we used a similar cost function with $c_{dist}^{run}=1.2$, $c_{dist}^{term}=1.2$, $c_{trace}^{run}=2.5$, $c_{col}^{run}=1,000$, with $\lambda=0.3$, $4096$ samples per step, $I_{2\times 2}$ MPPI control noise, and horizon of $H=8$. Since a binary collision term can degrade plan quality when many samples collide, we additionally included a finer-grained collision cost term penalizing known-obstacle penetration into the calibrated Gaussian confidence ellipse. This enables differentiating plans that are barely unsafe from those whose uncertainty ellipses deeply overlap with obstacles. This improved performance across methods.
\vspace{-3mm}
\section{Additional Results}\label{app:additionalResults}
\subsection{Planar Robot}\label{app:additionalResultsPlanar}
The test cases used for numerical coverage validation were collected by spanning the Cartesian grid given by
$(p_x,p_y)\in
\textsc{lin}(map_x^{\min},map_x^{\max},8)\allowbreak
\times\allowbreak
\textsc{lin}(map_y^{\min},\allowbreak map_y^{\max},8)$,
$(v_x,v_y) \in \textsc{lin}(-1.8,1.8,4)^2$,
and $(a_{x,t},a_{y,t}) \in \textsc{lin}(-0.9,0.9,4)^2$. Again, only non-colliding transitions were considered for both calibration dataset collection and test-case evaluation. Table \ref{tab:2dtestcasesFULL} expands Table \ref{tab:test2case2NOabblation} to include both ablations to \methodName. The ablation with access to test-environment data has a more relevant dataset, which is nevertheless smaller than our general $\calibset$. We observe similar performance to \methodName, which suggests that our perception information encoding enables an efficient calibration, even when using data from environments different from the evaluation environment. The ablation without perception information, Ablation w/o \textsc{Encode}($o^\prime_t$), can perform velocity-action-dependent calibration but cannot distinguish how uncertainty might vary between ID and OOD regions. Consequently, it is significantly over-conservative while ID, which is undesirable. This ablation's OOD coverage does not always meet the user-set requirement.

In Fig. \ref{fig:planarEnvInference} below, we compare the trajectories generated by each algorithm across the four tested environments (we ran 30 evaluations per environment-method combo). The trajectories of $\textit{LUCCa}$ and \methodName~are visually similar, even though our method \textit{does not require any data from the tested environment}. This is numerically supported by the metrics shown in Table \ref{tab:planning_2d_compact}. The ablation with access to test environment data produces results comparable to \methodName. It appears that using data from other environments (i.e., \methodName) might not necessarily lead to a noticeable performance drop, compared to using environment-specific data. This supports the \textit{utility of our approach for cross-environment local conformal calibration}. The ablation without perception information is visibly over-conservative in the ID regions, leading to longer average time-to-goal. No method got stuck due to high uncertainty.

\begin{table}[!t]
\vspace{-9mm}
\centering
\scriptsize
\setlength{\tabcolsep}{2pt}
\renewcommand{\arraystretch}{1.12}
\newsavebox{\testcasesDoubleIntegratorTableBox}
\begin{lrbox}{\testcasesDoubleIntegratorTableBox}
\color{black}
\begin{tabular}{c | l | c |
  cc @{\hspace{3pt}}!{\vrule width 0.3pt}@{\hspace{3pt}}
  cc @{\hspace{3pt}}!{\vrule width 0.3pt}@{\hspace{3pt}}
  cc @{\hspace{3pt}}!{\vrule width 0.3pt}@{\hspace{3pt}}
  cc}
\toprule
\multirow{2}{*}{Metric}
& \multicolumn{1}{c|}{\multirow{2}{*}{Method}}
& \multirow{2}{*}{\shortstack{Tested map\\\textbf{not} in $\calibset$?}}
& \multicolumn{2}{c@{\hspace{3pt}}!{\vrule width 0.3pt}@{\hspace{3pt}}}{Map S}
& \multicolumn{2}{c@{\hspace{3pt}}!{\vrule width 0.3pt}@{\hspace{3pt}}}{Map L}
& \multicolumn{2}{c@{\hspace{3pt}}!{\vrule width 0.3pt}@{\hspace{3pt}}}{Map H}
& \multicolumn{2}{c}{Map U} \\
& & & ID & OOD & ID & OOD & ID & OOD & ID & OOD \\
\midrule

\multirow{6}{*}{\shortstack{Marginal\\Coverage (\%)}}
& NoCP & N/A & \ccell{red!15}{89.9} & \ccell{red!15}{6.4} & \ccell{green!15}{90.0} & \ccell{red!15}{6.4} & \ccell{green!15}{90.0} & \ccell{red!15}{6.4} & \ccell{red!15}{89.9} & \ccell{red!15}{6.4} \\
& SplitCP & \myxmark & \ccell{green!15}{100.0} & \ccell{red!15}{69.7} & \ccell{green!15}{100.0} & \ccell{red!15}{69.1} & \ccell{green!15}{100.0} & \ccell{red!15}{61.8} & \ccell{green!15}{100.0} & \ccell{red!15}{71.6} \\
& LUCCa \cite{lucca} & \myxmark & \ccell{green!15}{91.4} & \ccell{green!15}{93.4} & \ccell{green!15}{91.1} & \ccell{green!15}{93.8} & \ccell{green!15}{91.0} & \ccell{green!15}{93.9} & \ccell{green!15}{90.6} & \ccell{green!15}{93.8} \\
& Ablation w/ test map data & \myxmark & \ccell{green!15}{91.5} & \ccell{green!15}{90.8} & \ccell{green!15}{90.4} & \ccell{green!15}{93.3} & \ccell{green!15}{90.5} & \ccell{green!15}{93.5} & \ccell{green!15}{92.6} & \ccell{green!15}{91.3} \\
& Ablation w/o \textsc{Encode}($o^\prime_t$) & \mycheck & \ccell{green!15}{100.0} & \ccell{red!15}{89.3} & \ccell{green!15}{100.0} & \ccell{green!15}{90.0} & \ccell{green!15}{100.0} & \ccell{green!15}{90.7} & \ccell{green!15}{100.0} & \ccell{red!15}{88.9} \\
& \textbf{OCULAR (ours)} & \mycheck & \ccell{green!15}{90.6} & \ccell{green!15}{93.4} & \ccell{green!15}{90.8} & \ccell{green!15}{93.7} & \ccell{green!15}{91.2} & \ccell{green!15}{91.0} & \ccell{green!15}{90.5} & \ccell{green!15}{93.8} \\
\midrule

\multirow{6}{*}{\shortstack{Median\\$\CPregion$ volume\\(wrt oracle) $\downarrow$}}
& NoCP & N/A & 0.99 & 0.02 & {1.00} & 0.02 & {1.00} & 0.02 & 0.99 & 0.02 \\
& SplitCP & \myxmark & 50.26 & 0.84 & 49.66 & 0.83 & 40.50 & 0.67 & 53.28 & 0.89 \\
& LUCCa \cite{lucca} & \myxmark & 1.11 & 1.14 & 1.06 & 1.14 & 1.09 & 1.11 & 1.06 & 1.14 \\
& Ablation w/ test map data & \myxmark & 1.07 & {1.12} & {1.00} & 1.10 & 1.05 & 1.11 & 1.07 & {1.11} \\
& Ablation w/o \textsc{Encode}($o^\prime_t$) & \mycheck & 57.86 & 0.97 & 59.37 & {0.99} & 60.56 & {1.01} & 58.38 & 0.98 \\
& \textbf{OCULAR (ours)} & \mycheck & {1.00} & {1.11} & 1.05 & 1.15 & {1.00} & 1.13 & {1.05} & 1.15 \\
\bottomrule
\end{tabular}
\end{lrbox}
\captionsetup{skip=4pt,width=\wd\testcasesDoubleIntegratorTableBox}
\caption{Test-case results across four planar environment maps.}
\label{tab:2dtestcasesFULL}
\color{black}
\centerline{\usebox{\testcasesDoubleIntegratorTableBox}}

\vspace{2pt}
\centerline{\begin{minipage}{\wd\testcasesDoubleIntegratorTableBox}
\scriptsize
\ifarxiv
\colorbox{red!15}{red}: coverage below $(1-\failureRate)=0.9$.
\else
Coverage $<90\%$: undercoverage.
\fi
$\CPregion$ volume reported as ratio relative to an oracle using the minimum scaling $\xi$ required to achieve $90\%$ coverage per transition.
Test transitions \#: S 4,096; L 2,816; H 4,096; U 6,656.
\end{minipage}}
\end{table}

\begin{table*}[!h]
\scriptsize
\setlength{\tabcolsep}{2pt}
\renewcommand{\arraystretch}{1.22}
\newsavebox{\planningDoubleIntegratorTableBox}
\begin{lrbox}{\planningDoubleIntegratorTableBox}
\color{black}
\begin{tabular}{c||c||c|c|c|c||c|c|c|c}
\toprule
\multirow{2}{*}{\centering Method}
& \multirow{2}{*}{\shortstack{Tested map\\\textbf{not} in $\calibset$?}}
& \multicolumn{4}{c||}{Success (\%) $\uparrow$}
& \multicolumn{4}{c}{Steps to completion (mean$\pm$std) $\downarrow$} \\
& & S & L & H & U & S & L & H & U \\
\midrule

\multicolumn{1}{l||}{NoCP}
& N/A
& 33.3 & 0 & 6.7 & 0
& \makebox[2.65em][r]{214.8}$\pm$\makebox[1.75em][l]{39.3} & -- & \makebox[2.65em][r]{{155.5}}$\pm$\makebox[1.75em][l]{{0.7}} & -- \\
\multicolumn{1}{l||}{SplitCP}
& \myxmark
& 0 & 0 & 0 & 0
& -- & -- & -- & -- \\
\multicolumn{1}{l||}{LUCCa \cite{lucca}}
& \myxmark
& {100} & {100} & {100} & {100}
& \makebox[2.65em][r]{{171.1}}$\pm$\makebox[1.75em][l]{{12.1}} & \makebox[2.65em][r]{{211.2}}$\pm$\makebox[1.75em][l]{{6.5}} & \makebox[2.65em][r]{{203.2}}$\pm$\makebox[1.75em][l]{{6.6}} & \makebox[2.65em][r]{{283.3}}$\pm$\makebox[1.75em][l]{{8.1}} \\
\multicolumn{1}{l||}{Ablation w/ test map data}
& \myxmark
& {100} & {100} & {100} & {100}
& \makebox[2.65em][r]{{170.9}}$\pm$\makebox[1.75em][l]{{5.3}} & \makebox[2.65em][r]{{206.7}}$\pm$\makebox[1.75em][l]{{6.4}} & \makebox[2.65em][r]{{206.0}}$\pm$\makebox[1.75em][l]{{8.5}} & \makebox[2.65em][r]{{280.3}}$\pm$\makebox[1.75em][l]{{6.8}} \\
\multicolumn{1}{l||}{Ablation w/o \textsc{Encode}($o^\prime_t$)}
& \mycheck
& {100} & {100} & {100} & {100}
& \makebox[2.65em][r]{181.7}$\pm$\makebox[1.75em][l]{4.7} & \makebox[2.65em][r]{229.1}$\pm$\makebox[1.75em][l]{10.0} & \makebox[2.65em][r]{{205.5}}$\pm$\makebox[1.75em][l]{{7.1}} & \makebox[2.65em][r]{293.6}$\pm$\makebox[1.75em][l]{8.4} \\
\multicolumn{1}{l||}{\textbf{OCULAR (ours)}}
& \mycheck
& {100} & {100} & {100} & {100}
& \makebox[2.65em][r]{{177.6}}$\pm$\makebox[1.75em][l]{{7.0}} & \makebox[2.65em][r]{{213.6}}$\pm$\makebox[1.75em][l]{{7.5}} & \makebox[2.65em][r]{{199.7}}$\pm$\makebox[1.75em][l]{{7.7}} & \makebox[2.65em][r]{{278.1}}$\pm$\makebox[1.75em][l]{{7.6}} \\
\bottomrule
\end{tabular}
\end{lrbox}
\captionsetup{skip=1pt,width=\wd\planningDoubleIntegratorTableBox}
\caption{Planning results across four planar environment maps ($30$ runs each).}
\label{tab:planning_2d_compact}

\color{black}
\centerline{
\usebox{\planningDoubleIntegratorTableBox}}
\par\vspace{3pt}
\noindent
\centerline{\begin{minipage}{\wd\planningDoubleIntegratorTableBox}
\scriptsize
Success $=$ reaching all subgoals without collisions.
\end{minipage}}
\end{table*}

\begin{figure}[!t]
\includegraphics[width=\columnwidth]{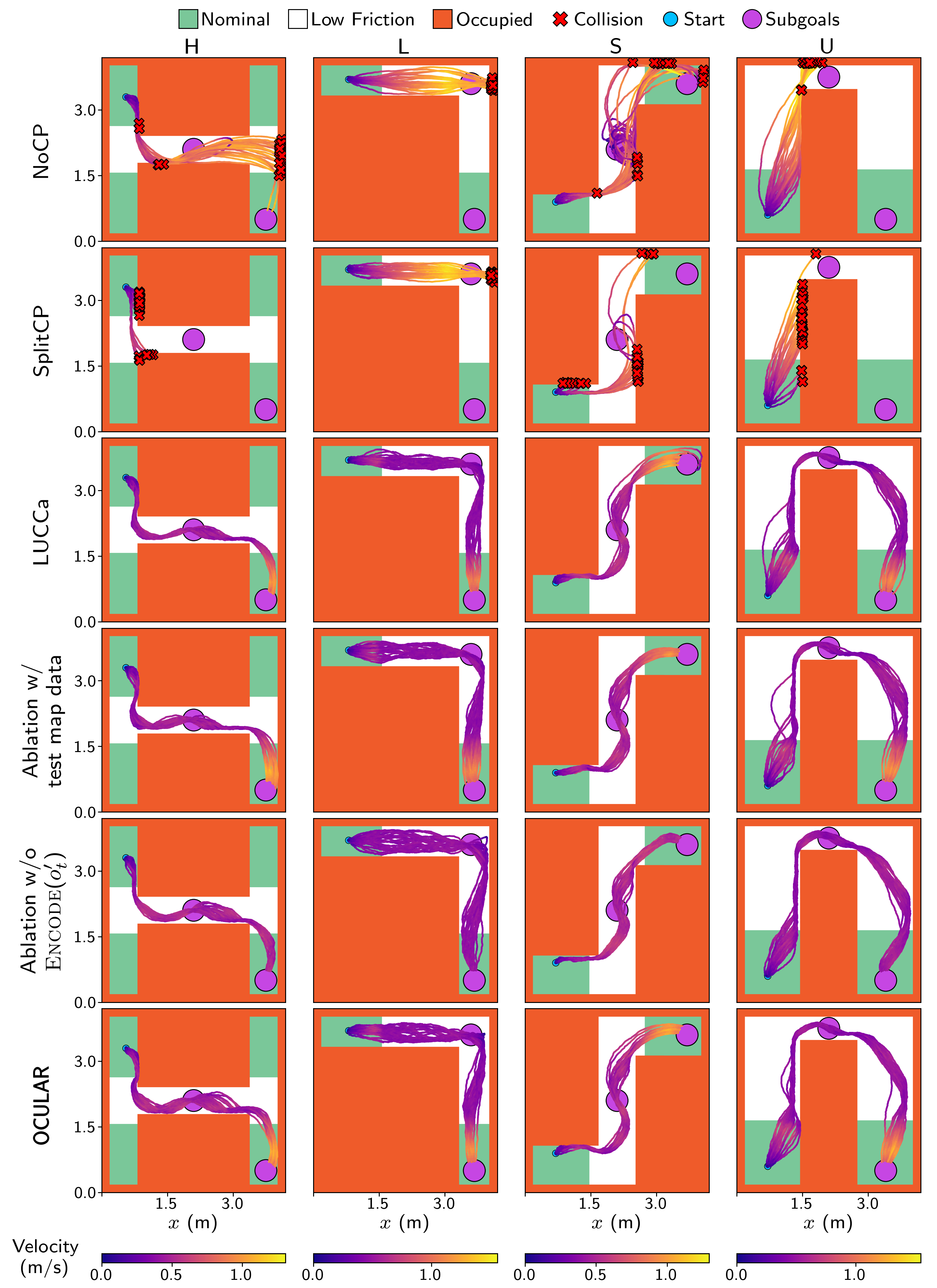}\vspace{-3mm}
\caption{\looseness-1 Trajectories of \methodName~and baselines in the planar environment, across 30 runs for each map-method combo. The uncalibrated baseline gains too much momentum in the low-friction region, leading to collisions. \textit{SplitCP} is too conservative, with all its sampled plans being in collision. This results in goal-chasing behavior and an inability to avoid obstacles. \textit{LUCCa} and \methodName~have comparable performance, slowing down in high-uncertainty regions, and reaching all subgoals without colliding. However, \textit{LUCCa} uses data specific to each tested map, while \textit{our method produces safe plans without any data from the executed map} (e.g., for map S: \textit{LUCCa} uses data collected in S; our method uses data from H, U, L). The ablation with access to test-time data performs comparably to \methodName, and the ablation without perception information is slower even in ID regions, moving equally slowly in both ID and OOD regions.} \label{fig:planarEnvInference}
\end{figure}
\FloatBarrier
\subsection{Isaac Sim}\label{app:additionalResultIsaac}
\vspace{-3mm}
\begin{figure}[b!]
\vspace{-4mm}
\includegraphics[width=\textwidth]{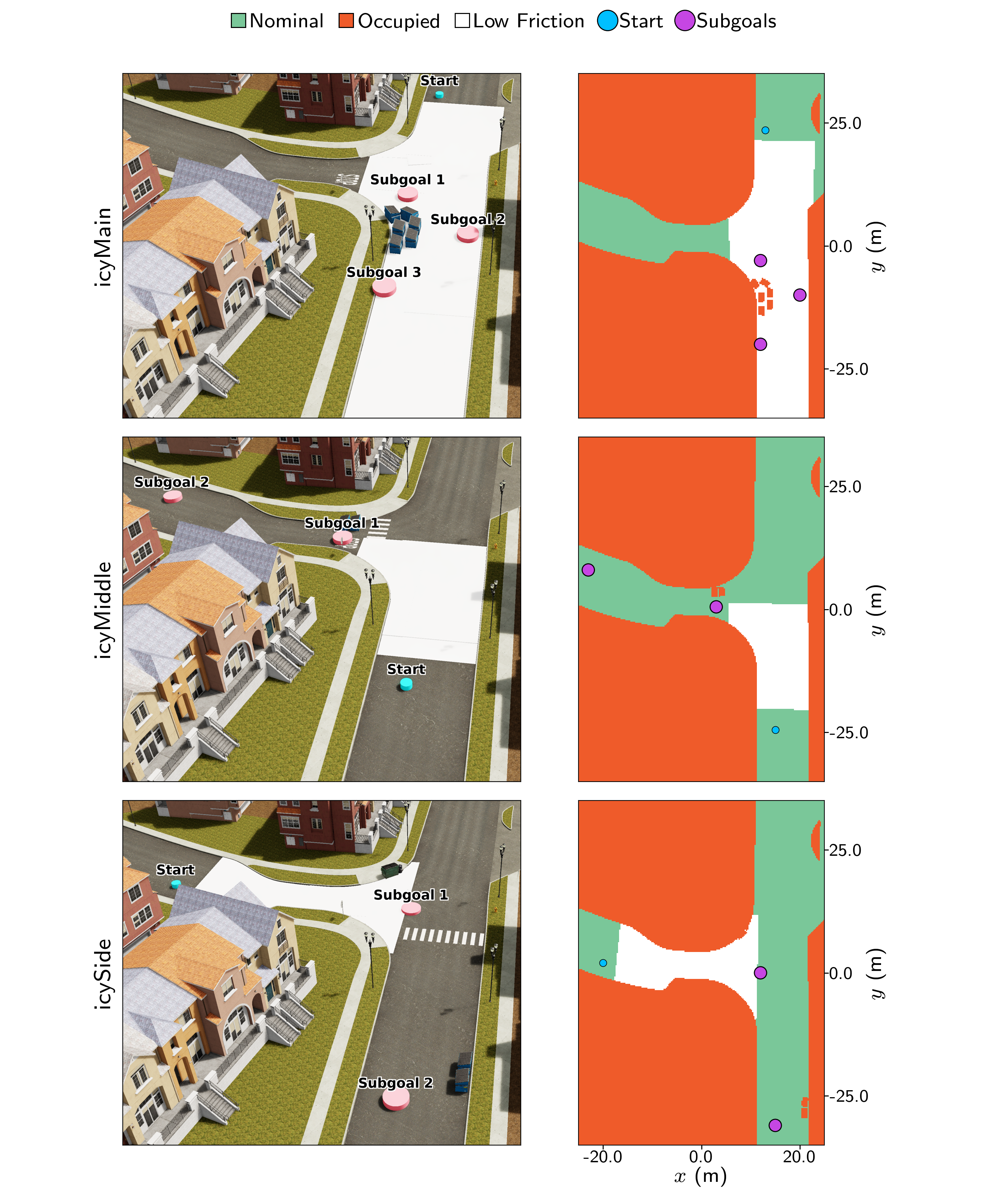}\vspace{-3mm}
\caption{\looseness-1  The three tested Isaac Sim environments (based off Rivermark).} \label{fig:isaacSimMaps}
\end{figure}
As in the planar experiment, calibration data was collected by spanning the Cartesian grid given by $(p_x,p_y) \in \textsc{lin}(map_x^{\min},map_x^{\max},16)\times \textsc{lin}(map_y^{\min},map_y^{\max},\allowbreak 16)$,
$(v_x,v_y) \in \textsc{lin}(-1.8,1.8,10)^2$,
and $(a_{x,t},a_{y,t}) \in \textsc{lin}(-0.9,0.9,4)^2$. Additionally, we excluded from the calibration and test data points that were close to the map edges and oriented outwards. This was done because our implementation does not maintain ground truth labels outside the map bounds. See Fig. \ref{fig:isaacSimMaps} for the three environments used. 

Fig. \ref{fig:isaacSensingExampleDumpsters} demonstrates an example observation and the perceived map after a single step (lighter color shades). This map is updated as new observations are received and used for trajectory optimization by all methods.

\begin{figure}[b!]
\includegraphics[width=\textwidth]{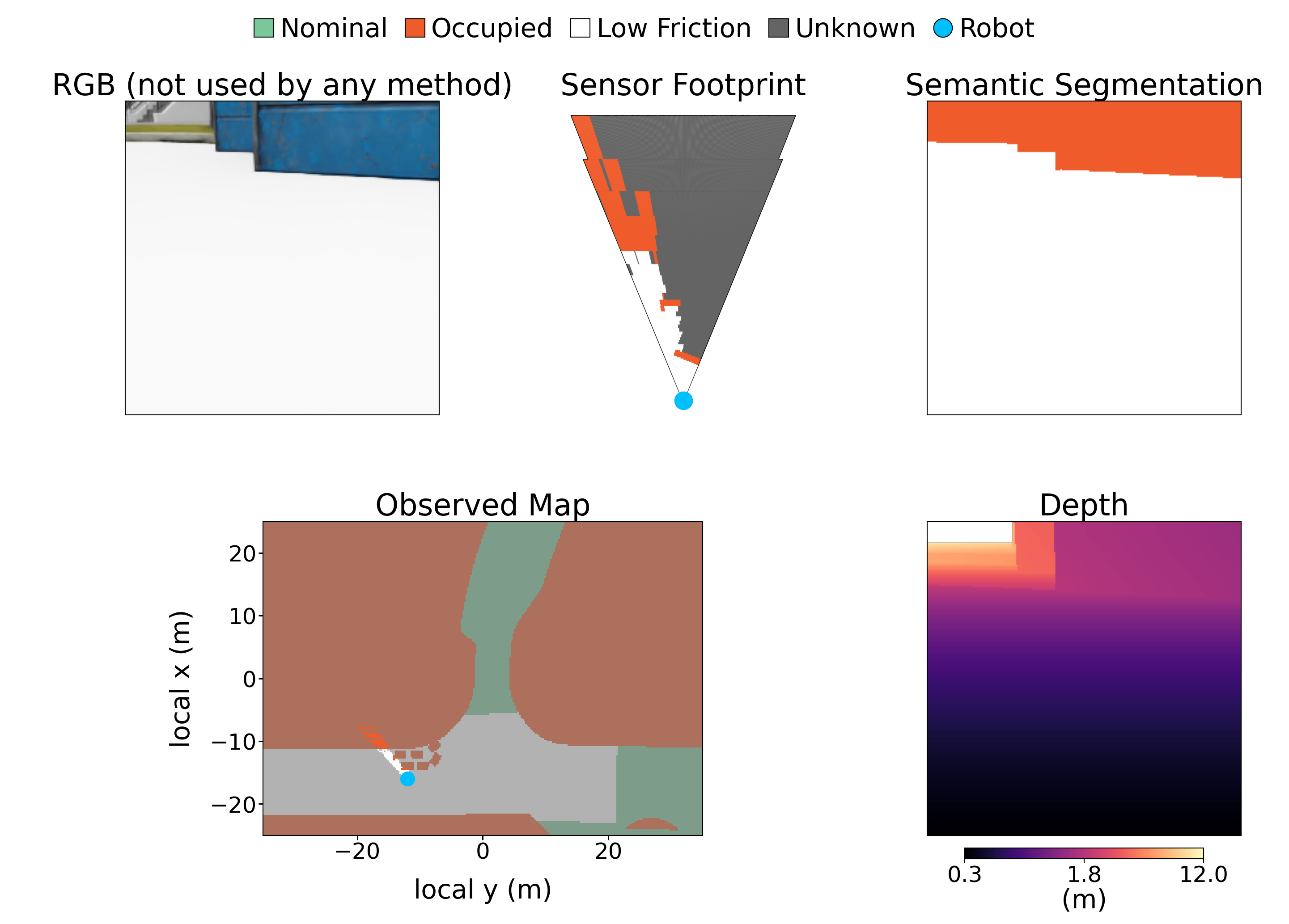}\vspace{-3mm}
\caption{\looseness-1 Example observation and observed map in the Isaac Sim environment. RGB is shown for clarity and is not used by any method. The depth and semantic segmentation images are converted into a planar sensor footprint, as detailed in Sec. \ref{sec:obsprocessing}. At the start of each episode, the planner starts from a fully unknown map. The darker shade cells represent the true underlying map semantics, which are unknown to the optimizer. Then, after capturing each observation, the planar semantic map is updated using $o_t^\prime$ as detailed in Sec.~\ref{sec:methodPlanning}. This map, represented by the lighter shade colors in the bottom left image, is then used for MPC planning at each step. Refer to Sec.~\ref{sec:methodPlanning} and the planning videos shown on the project webpage for more details.} \label{fig:isaacSensingExampleDumpsters}
\end{figure}

Table \ref{tab:isaacTestCasesFULL} expands on Table \ref{tab:isaacTestCases} by including the ablations. As in the planar setting, the ablation with access to test-time data has comparable performance to \methodName. The ablation without perception information, Ablation w/o \textsc{Encode}($o^\prime_t$), can considerably undercover when OOD and overcover when ID. This is unsurprising since, without access to observations, it does not have enough information to distinguish both regions. \newline\indent
Table \ref{tab:planning_isaac_compact_all_methods} reports the results from the safe motion planning experiments. The executed trajectories can be seen in Fig. \ref{fig:isaacEnvInference}. \textit{NoCP} gains significant momentum leading to collisions. \textit{SplitCP}, depending on the distribution between OOD and ID data in $\calibset$, can become over-conservative and get stuck in a narrow passage or become over-confident and gain unsafe amounts of speed. \textit{SplitCP} was the only method to time-out, i.e., take longer than 700 steps to reach all subgoals. This occurred on all its icySide runs and $20\%$ of its icyMiddle runs, with the remaining runs ending in collision.
Both \textit{LUCCa} and \methodName~appear to safely navigate all environments, slowing down as necessary to keep the next-state uncertainty manageable. The ablation with access only to test-time data performs similarly to our approach, being slightly faster or slower in some maps. This indicates that using cross-environment data might not lead to noticeable performance degradation compared with using environment-specific data. The ablation without perception information is overly conservative when OOD, as it cannot distinguish between nominal and low-friction regions. In icyMain, this ablation gains too much momentum leading to collisions. This indicates that access to perception information is important, as it could help determine how transition uncertainty changes across different environmental regions.

The test cases and planning results appear to indicate that our method is capable of producing \textit{volume-efficient} and \textit{adaptive} dynamics uncertainty calibrations, and consequently \textit{safe} motion plans when using more realistic perception information, even \textit{without access to any data from the execution environment}.

\begin{table*}[!h]
\vspace{-9mm}
\centering
\scriptsize
\setlength{\tabcolsep}{2pt}
\renewcommand{\arraystretch}{1.12}
\newsavebox{\testcasesPointCameraTableBox}
\begin{lrbox}{\testcasesPointCameraTableBox}
\begin{tabular}{c | l | c |
  cc @{\hspace{3pt}}!{\vrule width 0.3pt}@{\hspace{3pt}}
  cc @{\hspace{3pt}}!{\vrule width 0.3pt}@{\hspace{3pt}}
  cc}
\toprule
\multirow{2}{*}{Metric}
& \multicolumn{1}{c|}{\multirow{2}{*}{Method}}
& \multirow{2}{*}{\shortstack{Tested map\\\textbf{not} in $\calibset$?}}
& \multicolumn{2}{c@{\hspace{3pt}}!{\vrule width 0.3pt}@{\hspace{3pt}}}{icyMain}
& \multicolumn{2}{c@{\hspace{3pt}}!{\vrule width 0.3pt}@{\hspace{3pt}}}{icyMiddle}
& \multicolumn{2}{c}{icySide} \\
& & & ID & OOD & ID & OOD & ID & OOD \\
\midrule

\multirow{6}{*}{\shortstack{Marginal\\Coverage (\%)}}
& NoCP
& N/A
& \ccell{green!15}{90.0} & \ccell{red!15}{56.7}
& \ccell{green!15}{90.0} & \ccell{red!15}{56.7}
& \ccell{green!15}{90.0} & \ccell{red!15}{56.7} \\

& SplitCP
& \textcolor{red!65!black}{\myxmark}
& \ccell{green!15}{99.8} & \ccell{green!15}{93.0}
& \ccell{green!15}{99.1} & \ccell{red!15}{85.9}
& \ccell{green!15}{99.5} & \ccell{red!15}{89.6} \\

& LUCCa \cite{lucca}
& \textcolor{red!65!black}{\myxmark}
& \ccell{green!15}{90.1} & \ccell{green!15}{91.4}
& \ccell{green!15}{90.1} & \ccell{green!15}{90.9}
& \ccell{green!15}{91.1} & \ccell{green!15}{91.5} \\

& Ablation w/ test map data
& \textcolor{red!65!black}{\myxmark}
& \ccell{green!15}{92.2} & \ccell{green!15}{91.3}
& \ccell{green!15}{90.7} & \ccell{green!15}{90.2}
& \ccell{green!15}{91.3} & \ccell{green!15}{91.1} \\

& Ablation w/o \textsc{Encode}($o^\prime_t$)
& \textcolor{green!45!black}{\mycheck}
& \ccell{green!15}{96.0} & \ccell{red!15}{75.6}
& \ccell{green!15}{97.4} & \ccell{red!15}{82.1}
& \ccell{green!15}{97.2} & \ccell{red!15}{80.0} \\

& \textbf{OCULAR (ours)}
& \textcolor{green!45!black}{\mycheck}
& \ccell{green!15}{90.4} & \ccell{green!15}{91.0}
& \ccell{green!15}{91.1} & \ccell{green!15}{90.6}
& \ccell{green!15}{91.5} & \ccell{green!15}{90.1} \\
\midrule

\multirow{6}{*}{\shortstack{Median\\$\CPregion$ volume\\(wrt oracle) $\downarrow$}}
& NoCP
& N/A
& 1.00 & 0.28
& 1.00 & 0.28
& 1.00 & 0.28 \\

& SplitCP
& \textcolor{red!65!black}{\myxmark}
& 4.66 & 1.29
& 3.07 & 0.85
& 3.73 & 1.03 \\

& LUCCa \cite{lucca}
& \textcolor{red!65!black}{\myxmark}
& {1.02} & {1.10}
& {1.02} & 1.13
& 1.08 & 1.13 \\

& Ablation w/ test map data
& \textcolor{red!65!black}{\myxmark}
& 1.15 & 1.12
& 1.06 & {1.04}
& 1.08 & 1.11 \\

& Ablation w/o \textsc{Encode}($o^\prime_t$)
& \textcolor{green!45!black}{\mycheck}
& 1.88 & 0.54
& 2.39 & 0.71
& 2.28 & 0.65 \\

& \textbf{OCULAR (ours)}
& \textcolor{green!45!black}{\mycheck}
& {1.02} & 1.15
& 1.06 & 1.06
& {1.03} & {1.02} \\
\bottomrule
\end{tabular}
\end{lrbox}

\color{black}
\captionsetup{skip=4pt,width=\wd\testcasesPointCameraTableBox}
\caption{Test-case results across three Isaac Sim roads, including ablations.}
\label{tab:isaacTestCasesFULL}
\usebox{\testcasesPointCameraTableBox}

\begin{minipage}{\wd\testcasesPointCameraTableBox}
\scriptsize
\ifarxiv
\colorbox{red!15}{red}: coverage below $(1-\failureRate)=0.9$.
\else
Coverage $<90\%$: undercoverage.
\fi\ Volume reported as ratio relative to an oracle using the minimum $\xi$ to achieve $90\%$ coverage.
Test transition \#: icyMain 4,464; icyMiddle 4,464; icySide 4,464.
\end{minipage}
\vspace{-8mm}
\end{table*}

\begin{table*}[!b]
\vspace{-6mm}
\scriptsize
\setlength{\tabcolsep}{2pt}
\renewcommand{\arraystretch}{1.22}
\newsavebox{\planningPointCameraTableBox}
\begin{lrbox}{\planningPointCameraTableBox}
\color{black}
\begin{tabular}{c||c||c|c|c||c|c|c}
\toprule
\multirow{2}{*}{\centering Method}
& \multirow{2}{*}{\shortstack{Tested map\\\textbf{not} in $\calibset$?}}
& \multicolumn{3}{c||}{Success (\%) $\uparrow$}
& \multicolumn{3}{c}{Steps to completion (mean$\pm$std) $\downarrow$} \\
& & icyMain & icyMiddle & icySide & icyMain & icyMiddle & icySide \\
\midrule

\multicolumn{1}{l||}{NoCP}
& N/A
& 0 & 0 & 0
& -- & -- & -- \\

\multicolumn{1}{l||}{SplitCP}
& \textcolor{red!15!black}{\myxmark}
& 0 & 0 & 0
& -- & -- & -- \\

\multicolumn{1}{l||}{LUCCa \cite{lucca}}
& \textcolor{red!15!black}{\myxmark}
& {100} & {100} & {100}
& \makebox[2.65em][r]{332.1}$\pm$\makebox[1.75em][l]{13.1}
& \makebox[2.65em][r]{288.4}$\pm$\makebox[1.75em][l]{7.8}
& \makebox[2.65em][r]{339.1}$\pm$\makebox[1.75em][l]{8.7} \\

\multicolumn{1}{l||}{Ablation w/ test map data}
& \textcolor{red!15!black}{\myxmark}
& {100} & {100} & {100}
& \makebox[2.65em][r]{{280.0}}$\pm$\makebox[1.75em][l]{{6.1}}
& \makebox[2.65em][r]{305.9}$\pm$\makebox[1.75em][l]{28.5}
& \makebox[2.65em][r]{{211.1}}$\pm$\makebox[1.75em][l]{{5.9}} \\

\multicolumn{1}{l||}{Ablation w/o \textsc{Encode}($o^\prime_t$)}
& \textcolor{green!45!black}{\mycheck}
& 0 & {100} & {100}
& --
& \makebox[2.65em][r]{317.1}$\pm$\makebox[1.75em][l]{7.4}
& \makebox[2.65em][r]{282.3}$\pm$\makebox[1.75em][l]{4.9} \\

\multicolumn{1}{l||}{\textbf{OCULAR (ours)}}
& \textcolor{green!45!black}{\mycheck}
& {100} & {100} & {100}
& \makebox[2.65em][r]{311.3}$\pm$\makebox[1.75em][l]{4.7}
& \makebox[2.65em][r]{{278.6}}$\pm$\makebox[1.75em][l]{{6.4}}
& \makebox[2.65em][r]{{208.1}}$\pm$\makebox[1.75em][l]{{3.1}} \\
\bottomrule
\end{tabular}
\end{lrbox}

\captionsetup{skip=1pt,width=\wd\planningPointCameraTableBox}
\caption{Planning results across three Isaac Sim roads ($30$ runs each), including ablations.}
\label{tab:planning_isaac_compact_all_methods}

\centerline{\usebox{\planningPointCameraTableBox}}\par\vspace{3pt}
\noindent
\centerline{%
\begin{minipage}{\wd\planningPointCameraTableBox}
\scriptsize
Success $=$ reaching all subgoals without collisions.
\end{minipage}}
\end{table*}

\begin{figure}[t!]
\vspace{-3mm}
\includegraphics[width=\textwidth]{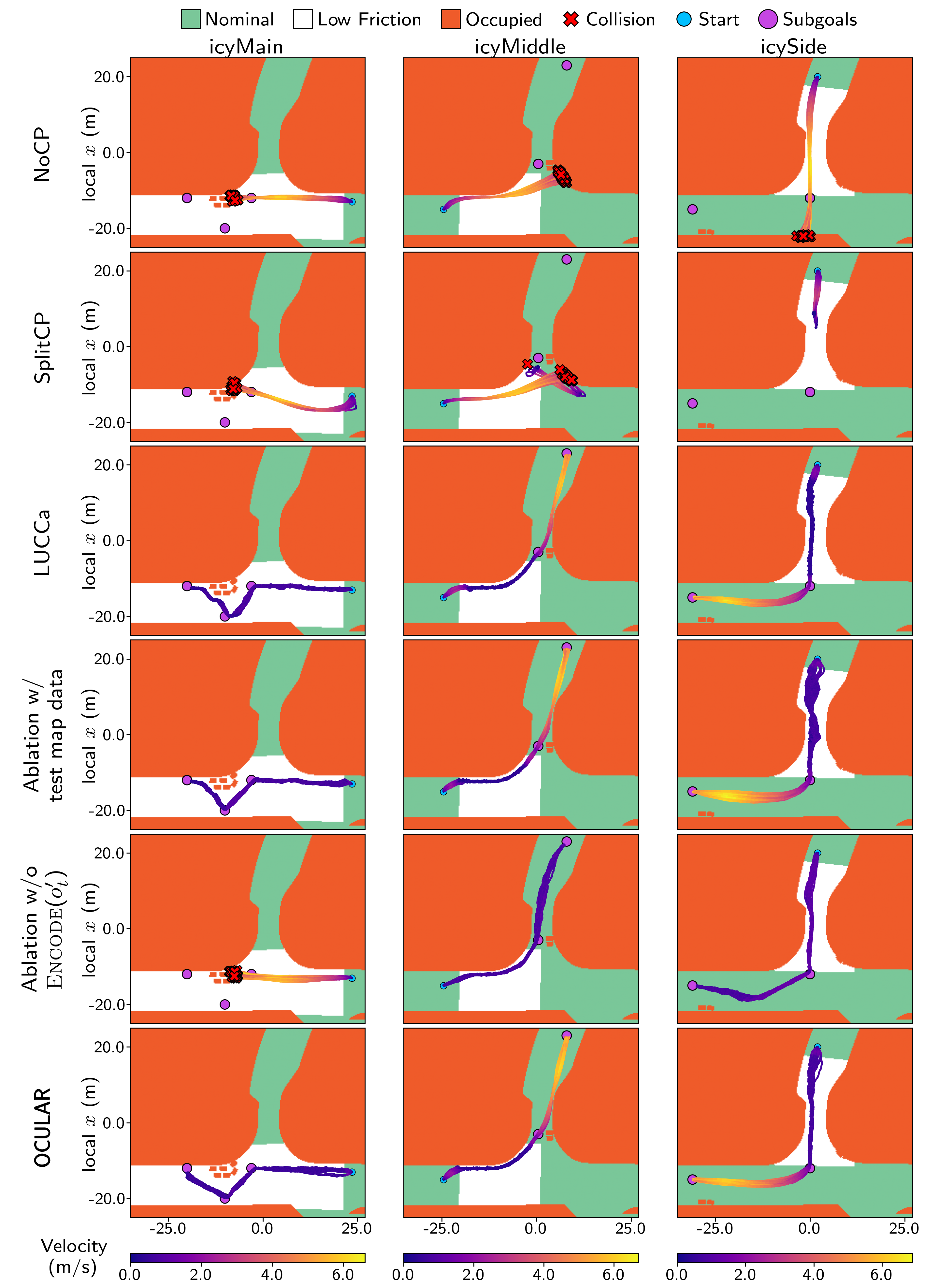}\vspace{-3mm}
\caption{\looseness-1  Comparison of all methods in the Isaac environment. \textit{NoCP} gains significant momentum when OOD, leading to collisions. \textit{SplitCP} can be over-conservative leading to time-outs in tight regions (e.g., icySide) or overconfident over ice (e.g., icyMain), since it cannot distinguish between observation-velocity-action inputs leading to lower or higher next-state uncertainty. \textit{LUCCa} and \methodName~have comparable performance, slowing down in high-uncertainty regions, and reaching subgoals safely. Yet, \textit{LUCCa} uses data specific to each tested environment, while our method produces safe and efficient plans \textit{without any data from the executed environment} (e.g., for map icyMain: \textit{LUCCa} uses data collected in icyMain; our method uses data collected from icyMiddle and icySide). The ablation with access to test-environment data performs comparably to our method. The ablation without perception information can become overly conservative over nominal terrain (icyMiddle, icySide) or overly optimistic when OOD (icyMain), as it cannot adapt to how the same actions can result in different amounts of uncertainty across regions.} \label{fig:isaacEnvInference}
\end{figure}

\fi

\end{document}